\documentclass[10pt]{article}

\usepackage[a4paper,margin=1in]{geometry}

\usepackage[T1]{fontenc}
\usepackage[utf8]{inputenc}
\usepackage{times}
\usepackage{microtype}

\usepackage{graphicx}
\usepackage{booktabs}
\usepackage{multirow}
\usepackage{wrapfig}
\usepackage{amsmath}
\usepackage{amssymb}
\usepackage{cite}
\usepackage{xspace}
\usepackage[dvipsnames]{xcolor}
\usepackage[hyphens]{url}
\usepackage{orcidlink}
\usepackage[accsupp]{axessibility}

\usepackage[labelfont=bf,font=small,tableposition=bottom]{caption}
\usepackage[skip=3pt]{subcaption}

\definecolor{linkblue}{rgb}{0.20,0.45,0.95}
\usepackage{hyperref}
\hypersetup{
    breaklinks=true,
    colorlinks=true,
    allcolors=linkblue,
    pdfborder={0 0 0},
    pdfhighlight=/N
}

\usepackage[capitalize]{cleveref}
\crefname{section}{Sec.}{Secs.}
\Crefname{section}{Section}{Sections}
\crefname{table}{Tab.}{Tabs.}
\Crefname{table}{Table}{Tables}

\definecolor{LightCyan}{rgb}{0.88,1,0.88}
\definecolor{LightCyan2}{rgb}{0.0,0.7,0.0}
\definecolor{DarkGreen}{rgb}{0.0,0.0,0.7}
\definecolor{bananamania}{rgb}{0.98, 0.91, 0.71}

\usepackage{algorithm}      % floats the algorithm
\usepackage{algpseudocode}  % provides the algorithmic environment
\usepackage[table]{xcolor}
\usepackage{comment}

\title{Neural Collapse-Inspired Multi-Label Federated Learning under Label-Distribution Skew}

\author{
Can Peng$^{1}$, 
Yuyuan Liu$^{1}$, 
Yingyu Yang$^{1}$, 
Pramit Saha$^{1}$, 
Qianye Yang$^{1}$, 
and J.\ Alison Noble$^{1}$ \\[0.5em]
$^{1}$University of Oxford, Oxford, United Kingdom
}

\date{}

\begin{document}

\maketitle

\begin{abstract}
Federated Learning (FL) enables collaborative model training across distributed clients while preserving data privacy, but remains challenging when client data are highly heterogeneous.
These challenges are further amplified in multi-label scenarios, where inter-label dependencies and mismatches between local and global label relationships introduce additional optimization conflicts.
While most FL studies focus on single-label classification, many real-world applications are inherently multi-label and often exhibit severe label skew across clients.
To address this important yet underexplored problem, we propose FedNCA-ML, a novel FL framework that aligns client representations and learns discriminative, well-clustered features inspired by Neural Collapse (NC) theory.
NC describes an ideal latent geometry where each class’s features collapse to their mean, forming a maximally separated simplex.
FedNCA-ML further introduces an attention-based module to extract class-specific representations, enabling more balanced learning under heavy label imbalance.
These class-wise representations are then aligned via a shared NC-inspired structure, mitigating inter-client conflicts induced by heterogeneous local data and inconsistent label dependencies.
In addition, we design regularisation losses to encourage compact and consistent feature clustering in the latent space.
Experiments on five benchmark datasets under nine FL settings demonstrate the effectiveness of the proposed method, achieving improvements of up to 3.92\% in class-wise AUC and 4.93\% in class-wise F1 score.
\end{abstract}

\section{Introduction}
\label{sec: introduction}

\begin{figure}[t]
    \centering
    \includegraphics[width=1\linewidth, trim={0.6cm 0.1cm 0.4cm 0.1cm},clip]{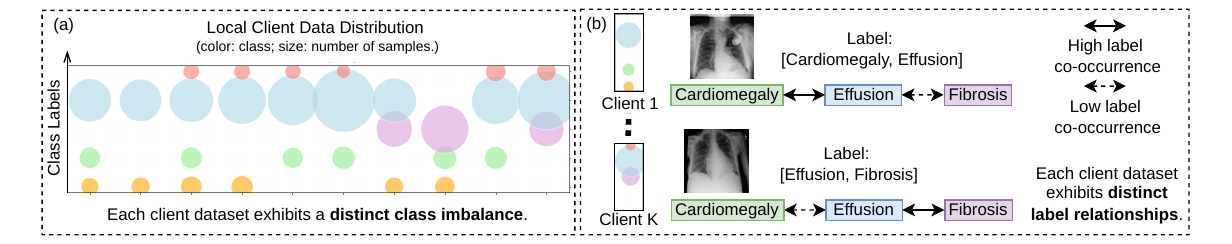} % left bottom right top
    \caption{
    Multi-label, label-skewed federated learning poses three key challenges:
    (1) heterogeneous client data with distinct class-imbalance patterns,
    (2) amplified heterogeneity due to multi-label co-occurrence structure,
    and (3) cross-client inconsistency in both data distributions and label relationships.
    }
	\label{fig1: abstract}
\end{figure}

Federated Learning (FL) enables collaborative model training on sensitive data, particularly in medical imaging, where privacy regulations and legal constraints prohibit data sharing.
However, most existing FL methods primarily target standard multi-class classification and overlook more realistic scenarios in which multiple objects or conditions co-occur within a single sample. 
For instance, many diseases are interrelated, and patients frequently present with multiple concurrent conditions.
Meanwhile, each client (e.g., hospital) may have a uniquely skewed data distribution due to variations in geography, population demographics, medical facilities, and clinical expertise.
In this paper, we study multi-label FL under label-skew settings, where each client’s local data is highly imbalanced and can substantially deviate from the global distribution. 
Figure~\ref{fig1: abstract} illustrates this scenario, showing that clients possess heterogeneous data distributions and distinct label dependency patterns. 
Without sharing raw data, our goal is to collaboratively train a global model that performs well across all target classes.

The multi-label, label-skewed FL setting poses three key challenges. 
\textbf{(1) Imbalanced data distribution.}
Local datasets often exhibit severe label imbalance, containing majority, minority, or even missing classes. 
Such skew drives each client to optimize toward its local distribution, typically overfitting dominant labels while under-training rare ones.
Consequently, clients learn locally biased features that mainly benefit their own data, leading to a global model with impaired generalization.
Moreover, imbalance is not limited to individual clients, since the overall label distribution is often imbalanced as well, further amplifying the difficulty of balanced learning.
\textbf{(2) Multi-label co-occurrence bias.}
Multi-label data further exacerbates the imbalance due to label co-occurrence. Frequent labels often appear alongside many others and dominate the training signal, suppressing the learning of discriminative features for minority labels and making rare conditions harder to recognize.
\textbf{(3) Cross-client inconsistency in label distributions and relationships.}
Under FL, clients differ in both label frequencies and label dependency structures.
These mismatches lead to optimization conflicts across clients, complicating collaborative training and hindering global convergence and generalization.

To address these challenges, we propose \underline{Fed}erated \underline{N}eural \underline{C}ollapse \underline{A}lignment for \underline{M}ulti-\underline{L}abel Learning (\textbf{FedNCA-ML}). 
FedNCA-ML is a unified representation learning framework for multi-label FL, inspired by Neural Collapse (NC) theory~\cite{papyan2020prevalence}, that promotes a consistent and discriminative feature geometry across heterogeneous clients. 
It is designed with two goals in mind. 
First, it encourages the model to learn balanced, class-discriminative representations, so that tail labels receive comparable attention to head labels. 
Second, it mitigates client drift by guiding local training toward a shared global geometry, reducing the tendency of each client to over-specialize to its own data distribution.
NC theory provides a geometric lens for this purpose. 
It shows that, when a classifier is trained to saturation on a balanced multi-class dataset, class-mean features converge to a simplex configuration and align with an Equiangular Tight Frame (ETF). 
This provides a principled geometric prior and has motivated a growing line of NC-inspired methods that encourage such structured representations under non-ideal training conditions and downstream tasks~\cite{li2023no,yang2023neural,gao2024distribution}.
FedNCA-ML introduces a shared global ETF prior to promote client-agnostic clustering and align local models within a common feature space that supports all classes, thereby reducing overfitting to client-specific biases.

Furthermore, the NC theory was originally developed for single-label classification, where each image representation corresponds to exactly one class.
In multi-label classification, a single shared representation is often insufficient to capture class-specific evidence and complex label relationships.
FedNCA-ML introduces an attention-based module that extracts class-wise representations from the shared image features, effectively reformulating multi-label learning into a set of per-class subproblems compatible with NC-style alignment. 
Consequently, ETF anchoring is applied only to these class-wise representations, rather than enforcing mutual exclusivity on the shared backbone space, allowing semantic proximity to remain naturally preserved in the backbone features.
Finally, two complementary regularizers further improve compactness and robustness.
A rejection loss suppresses noisy negative features, while a contrastive loss promotes tight intra-class clustering and clear inter-class separation.
The key contributions of this paper are as follows:
\begin{itemize}
\item We formalize the problem of multi-label FL under label skew, where clients differ in both label frequencies and label co-occurrence patterns.
\item We propose FedNCA-ML, an NC–inspired representation alignment framework that enforces a shared ETF geometry across clients to mitigate representation drift and improve balanced learning.
\item We introduce a class-wise attention mechanism that enables NC alignment in multi-label settings while preserving semantic relationships in the shared backbone feature space.
\item We introduce complementary rejection and contrastive regularizers that enhance intra-class compactness and inter-class separation under heterogeneous label distributions.
\end{itemize}

\section{Related Work}
\label{sec: related_work}

\noindent\textbf{Heterogeneous Federated Learning.}
FedAvg~\cite{mcmahan2017communication}, which iteratively aggregates client models via weighted parameter averaging, remains the foundation of most FL methods.
However, its performance often degrades under non-IID data.
Such heterogeneity typically arises from quantity skew, label distribution skew, and feature distribution skew.
To mitigate these issues in multi-class classification, many FL methods have been proposed.
Existing methods generally address heterogeneity from three perspectives.
\textbf{(1) Local learning phase.}
These methods steer client-side optimization by incorporating regularization terms~\cite{li2020federated, shoham2019overcoming, zhang2021federated} and/or auxiliary objectives~\cite{guo2025exploring, li2021fedrs, li2021model, wu2023fediic} to reduce the drift between local updates and the global model.
\textbf{(2) Post-learning phase.}
These methods correct the divergence among local models at the server during aggregation, either by exchanging additional information~\cite {karimireddy2020scaffold} or by using more robust aggregation strategies~\cite{shen2021agnostic}.
\textbf{(3) Pre-learning phase.}
This direction predefines a shared, fixed classifier or decision boundary to align local training, encouraging more consistent representation learning without modifying the core optimization procedure~\cite{dong2022spherefed, li2023no}.
In this work, we study multi-label FL under quantity-skewed and label-skewed distributions. 
Directly extending existing strategies to multi-label FL is often less effective, particularly for pre-learning methods, because multiple co-occurring labels per sample make it difficult to enforce clear semantic clustering and class separation in the latent space.
To cope with severe class imbalance, missing labels on some clients, and heterogeneous label co-occurrence patterns, we propose a pre-learning approach that explicitly organizes the latent feature space, improving robustness and generalization under label-skewed multi-label FL.

\noindent\textbf{Multi-Label Federated Learning.}
Multi-label FL poses unique challenges beyond single-label FL due to heterogeneous label co-occurrence across clients.
FedMLP~\cite{sun2024fedmlp} targets partial class annotation under task heterogeneity, where each client observes only a subset of labels with incomplete annotations.
To alleviate missing annotations, it exchanges local model parameters and class-wise prototypes with the server to enable pseudo-labeling.
FedLGT~\cite{liu2024language} studies a closely related setting and leverages label text to model shared label structure across clients.
Building on C-Tran~\cite{lanchantin2021general}, it adopts frozen CLIP~\cite{radford2021learning} text embeddings to enforce a consistent label-embedding space, thereby reducing cross-client divergence in label dependencies.

\noindent\textbf{Neural Collapse-Inspired Methods.}
Recent studies show that overparameterized networks trained to saturation on balanced multi-class datasets exhibit Neural Collapse (NC) in the terminal phase of training (TPT)~\cite{papyan2020prevalence}.
In this regime, last-layer features concentrate around their class means, and the class prototypes converge to an Equiangular Tight Frame (ETF), yielding a highly symmetric and well-separated latent geometry.
This observation has motivated NC-inspired methods that explicitly encourage such structure across a range of tasks~\cite{li2023no, yang2023neural, wei2025compress, gao2024distribution, liu2023inducing}.
However, NC has been studied mainly in multi-class classification, while multi-label settings remain less explored.
Li~\emph{et al.}~\cite{li2023neural} reported a generalized NC behavior in balanced multi-label training that features from single-label samples still follow a simplex ETF structure, whereas multi-label features are well approximated as scaled averages of single-label prototypes.
MLC-NC~\cite{tao2025mlc} exploits NC to improve long-tailed multi-label classification in a centralized setting.
To the best of our knowledge, we are the first to investigate multi-label FL through the lens of NC, leveraging it to mitigate interference induced by heterogeneous client distributions and label relationships.
\section{Preliminaries}
\label{sec: Preliminaries}

\subsection{Problem Formulation} 
We consider a multi-label, label-skewed FL setting with \(K\) clients collaboratively training a shared global model. 
Each client \(k\) holds a private dataset \(\mathcal{D}_k=\{(x_i^k,y_i^k)\}_{i=1}^{N_k}\) of size \(N_k\), where the input image \(x_i^k\in\mathbb{R}^{H\times W\times 3}\) and the label \(y_i^k\in\{0,1\}^{C}\) is a multi-hot vector over the \(C\) classes.
Let \(\mathcal{C}=\{1,\dots,C\}\) denote the global class index set.
Due to label skew, client \(k\) only observes a subset \(\mathcal{C}_k\subseteq\mathcal{C}\), and even for shared classes, the class-conditional distributions can differ across clients.
The goal is to learn a single global model that accurately recognizes all classes in \(\mathcal{C}\) without sharing any local client data.

\subsection{Neural Collapse (NC)}
In this section, we first introduce the structure of the simplex ETF, followed by a discussion of the key properties of NC~\cite{papyan2020prevalence, li2023neural}.

\noindent\textbf{Simplex Equiangular Tight Frame (ETF).}
A simplex ETF matrix \(\mathbf{M} = [\mathbf{m}_c]_{c=1}^{C} \in \mathbb{R}^{d\times C}\) composed of \(C\) column vectors, each corresponding to a class prototype in \(\mathbf{m}_c\in\mathbb{R}^{d}\).
A standard construction is:
\begin{equation}
\mathbf{M}
\;=\;
\sqrt{\frac{C}{C-1}}\;\mathbf{U}\;\Big(\mathbf{I}_C - \frac{1}{C}\,\mathbf{1}_C\mathbf{1}_C^{\top}\Big),
\label{eq:etf-construct}
\end{equation}
where \(\mathbf{U}\in\mathbb{R}^{d\times C}\) denotes a rotation orthogonal matrix \((\mathbf{U}^{\top}\mathbf{U}=\mathbf{I}_C)\). 
\(\mathbf{I}_C\) is the \(C\times C\) identity matrix.
\(\mathbf{1}_C\) is the \(C\)-dimensional all-ones vector.

This yields unit-norm columns with equal pairwise inner products:
\begin{equation}
\mathbf{m}_a^{\top}\mathbf{m}_b
=
\begin{cases}
1, & a=b,\\[2pt]
-\dfrac{1}{C-1}, & a\neq b,
\end{cases}
\qquad a,b\in\mathcal{C}.
\label{eq:etf-inner}
\end{equation}
Hence, all prototypes have equal \(\ell_2\) norm and identical pairwise angles, forming a centered regular simplex.

\noindent\textbf{Neural Collapse properties.}
At the end of training, networks exhibit the NC phenomenon. 
Empirically, the following regularities are observed:

\noindent\textbf{ \(\mathcal{NC}_1\): Variability Collapse.} 
Within-class feature variance collapses, causing features of the same class to concentrate around their class mean.

\noindent\textbf{ \(\mathcal{NC}_2\): Convergence to Simplex ETF.} 
Class means arrange themselves as the vertices of a simplex ETF, forming a maximally symmetric and equidistant configuration.

\noindent\textbf{ \(\mathcal{NC}_3\): Self-Duality.}
Upon appropriate rescaling, the final-layer classifier weights align with the class means, exhibiting the same simplex ETF geometry.

\begin{figure}[t]
    \centering
    \includegraphics[width=1\linewidth, trim={0.5cm 0.2cm 0.5cm 0cm},clip]{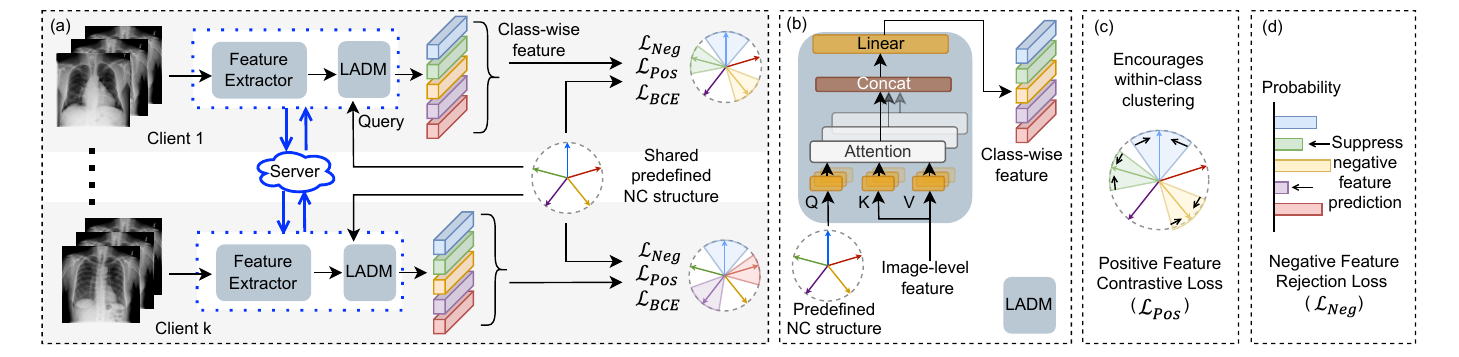} % left bottom right top
    \caption{
    Overview of the proposed FedNCA-ML framework for multi-label label-skewed FL.
    Subfigure (a) shows the overall architecture, while Subfigures (b)–(d) illustrate the Label-Aware Disentanglement Module (LADM) and the regularization losses.
    The attention-based LADM extracts label-specific features from image-level features.
    A predefined ETF matrix acts as both the shared classifier and the source of class-wise query embeddings, ensuring consistent local training across clients. 
    Two regularisation terms are further incorporated to suppress noisy negative features and promote compact intra-class clustering in the latent feature space.
    }
	\label{fig: framework}
\end{figure}

\section{Proposed Method}
\label{sec: Proposed Method}
In multi-label, label-skewed FL, each client’s data is highly imbalanced and contains missing labels. 
Clients can also exhibit distinct label co-occurrence patterns, and both label distributions and dependencies may vary widely due to heterogeneous data collection. 
Together, these factors amplify client drift and hinder stable convergence during local training and aggregation.
To address these challenges, we propose \textbf{FedNCA-ML}, which leverages an NC-inspired simplex ETF as a shared feature-space reference to encourage balanced learning across classes and provide a consistent decision boundary for aligning local training. 
An overview is shown in Fig.~\ref{fig: framework}.
FedNCA-ML consists of three key components.
(1) We employ an attention-based module (LADM; Fig.~2b) to extract class-wise features from the backbone’s shared image-level representation, and apply ETF anchoring only to these class-wise features (Fig.~2a). 
This avoids directly constraining the shared image-level representation, allowing semantic proximity to be naturally preserved in the backbone features. 
(2) We use a predefined ETF as a globally fixed classifier, providing a consistent alignment target across clients (Fig.~2a). 
(3) We introduce two additional regularisers (Fig.~2c--d) to further improve clustering in the latent space.

\subsection{Label-Aware Disentanglement Module}
In principle, a single pooled embedding can encode multiple labels. 
However, under severe label imbalance, it often entangles class-specific evidence, inducing gradient interference and bias toward majority labels.
Motivated by disentanglement-based multi-label learning, we introduce a Label-Aware Disentanglement Module (LADM) to extract class-wise features from backbone representations~\cite{liu2021query2label,ridnik2023ml,xu2023label}.
LADM is further inspired by an analogy to object detection that overlapping bounding boxes indicate that the same region may support multiple instances.
Similarly, in multi-label recognition, a single region can provide evidence for multiple semantic categories, particularly when labels co-occur.
Accordingly, LADM adopts a DETR-style cross-attention mechanism~\cite{carion2020end}.

Concretely, a set of class queries attends to a grid of image tokens to produce class-specific representations.
Unlike DETR, where queries represent instances and are trained with box-level supervision, LADM assigns one fixed query to each class and learns solely from image-level annotations.
Each query acts as a soft region selector, aggregating spatial evidence relevant to its class while leveraging contextual information across regions.
This yields a set of disentangled, class-aware feature vectors for per-class prediction.
In FL, non-IID local data can induce inconsistent class-wise feature extraction across clients, destabilizing aggregation.
To encourage global consistency, LADM shares a single query matrix across all clients, anchoring each class to the same query direction.
This shared design reduces inter-client conflicts during aggregation, and its effectiveness is validated in the ablation study (Section~\ref{sec: ablation_study}, Table~\ref{tab: Ablation_LADM_Derma-MNIST}).

Formally, for client \(k\), each sample \(x_i^k \in \mathcal{D}_k\) (\(i \in [N_k]\)) is encoded by the backbone into a spatial feature map:
\begin{equation}
\mathbf{F}_i^k \in \mathbb{R}^{d \times H' \times W'},
\label{eq:LADM_F}
\end{equation}
where \(d\) denotes the channel dimension and \(H'\times W'\) is the feature resolution.
The feature map is then flattened into a sequence of \(S = H' W'\) tokens:
\begin{equation}
\mathbf{Z}_i^k = \operatorname{reshape}(\mathbf{F}_i^k)^\top \in \mathbb{R}^{S \times d},
\label{eq:LADM_reshape}
\end{equation}
and, for notational convenience, we omit the client index in subsequent equations and denote the token sequence as \(\mathbf{Z}_i\).
To preserve spatial structure,  we add a fixed 2D sine–cosine positional embedding~\cite{carion2020end} 
to the key and value projections:
\begin{equation}
\tilde{\mathbf{Z}}_i = \mathbf{Z}_i + \mathbf{P},
\quad 
\mathbf{P} = \mathrm{PE}_{2\mathrm{D}}(H', W', d),
\label{eq:LADM_position}
\end{equation}
where $\mathrm{PE}_{2\mathrm{D}}(\cdot)$ denotes the fixed sine–cosine positional encoding. 
This embedding introduces spatial awareness without adding learnable parameters, thereby ensuring a consistent inductive bias across clients.

LADM employs a 4-head cross-attention to obtain class-specific features from the encoded image tokens.
We define a shared simplex ETF matrix as
\begin{equation}
\mathbf{M} = [\mathbf{m}_c]_{c=1}^{C} \in \mathbb{R}^{d\times C},
\label{eq:LADM_ETF}
\end{equation}
where each column \(\mathbf{m}_c\) serves both as a fixed classifier weight and as a class-specific query vector.  
Given the query \(\mathbf{m}_c\) and the encoded image tokens \(\tilde{\mathbf{Z}}_i\), LADM computes the class-specific feature using multi-head cross-attention:
\begin{equation}
\mathbf{h}_{ic} \;=\;
\operatorname{MultiHeadAttn}
\bigl(\mathbf{m}_c^{\top},\, \tilde{\mathbf{Z}}_i,\, \tilde{\mathbf{Z}}_i\bigr),
\quad c \in \mathcal{C}.
\label{eq:LADM_h}
\end{equation}
The resulting features are stacked to form the class-feature matrix:
\begin{equation}
\mathbf{H}_i \;=\;
\begin{bmatrix}
\mathbf{h}_{i1}^{\top}\\[-2pt]\vdots\\[-2pt]\mathbf{h}_{iC}^{\top}
\end{bmatrix}
\in\mathbb{R}^{C\times d}.
\label{eq:LADM_H}
\end{equation}

\subsection{Neural Collapse-Inspired Feature Alignment} 
After obtaining class-wise features with LADM, each sample \(i \in [N_k]\) yields a feature \(\mathbf{h}_{ic}\in\mathbb{R}^d\) for every class \(c \in \mathcal{C}\).
To prevent client-specific classifier drift, we fix the classifier to a globally shared simplex ETF, which also serves as the LADM query matrix.
This shared classifier enhances class separability and mitigates model drift caused by class imbalance and missing labels.
Given \(\mathbf{h}_{ic}\) and its corresponding prototype \(\mathbf{m}_c\), we compute the class logit and apply the sigmoid function to obtain a binary prediction:
\begin{equation}
\hat{y}_{ic} \;=\; \sigma\!\big(\mathbf{h}_{ic}^{\top}\mathbf{m}_{c}\big),
\label{eq:sigmoid}
\end{equation}
and train with a binary cross-entropy (BCE) loss:
\begin{equation}
\mathcal{L}_{\text{BCE}}
= -\frac{1}{N}\frac{1}{C}\sum_{i=1}^{N}\sum_{c=1}^{C}\Big[
y_{ic}\,\log\hat{y}_{ic} + (1-y_{ic})\,\log(1-\hat{y}_{ic})
\Big].
\label{eq:bce}
\end{equation}
\noindent Here, \(y_{ic}\in\{0,1\}\) denotes the ground-truth label indicator, and \(\sigma(\cdot)\) is the sigmoid function.

\subsection{Reducing Noise and Improving Clustering}
Each sample produces \(C\) class-wise features \(\{\mathbf{h}_{ic}\}_{c=1}^C\).
Let the positive and negative class sets for sample \(i\) be defined as
\(
\mathcal{C}_i^{+}=\{\,c\mid y_{ic}=1\,\}
\)
and
\(
\mathcal{C}_i^{-}=\{\,c\mid y_{ic}=0\,\}
\),
respectively.
We introduce two regularization terms: one to suppress noise from negative features, and another to promote compact clustering of positive features.

\noindent\textbf{Negative Feature Rejection Loss.}
While \(\mathcal{L}_{\text{BCE}}\) discourages alignment between a negative feature \(\mathbf{h}_{ic}\) and its corresponding prototype \(\mathbf{m}_c\) when \(y_{ic}=0\), it does not prevent \(\mathbf{h}_{ic}\) from spuriously aligning with other class prototypes. 
To address this, we introduce a penalty on high similarity between each negative feature and all non-self prototypes:
\begin{equation}
\hat{s}_{icr} \;=\; \sigma\!\big(\mathbf{h}_{ic}^{\top}\mathbf{m}_{r}\big),
\quad c\in\mathcal{C}_i^{-},\; r\in\mathcal{C}\setminus\{c\},
\label{eq:neg}
\end{equation}
and define the negative feature rejection loss as
\begin{equation}
\mathcal{L}_{\text{Neg}}
= -\frac{1}{N}\sum_{i=1}^{N}\frac{1}{|\mathcal{C}_i^{-}|}\sum_{c\in\mathcal{C}_i^{-}}\frac{1}{C-1}\sum_{\substack{r=1\\ r\neq c}}^{C}
\mathbb{I}\!\big(\hat{s}_{icr}>\tau\big)\,\log\!\big(1-\hat{s}_{icr}\big)
\label{eq:neg-loss}
\end{equation}
\noindent The indicator \(\mathbb{I}(\cdot)\) filters out low-similarity pairs, ensuring that only confident negatives contribute to the loss.
In our experiments, we set \(\tau=0.3\).

\noindent\textbf{Positive Feature Contrastive Loss.}
To encourage compact and discriminative class-wise clustering in the latent feature space, we introduce a contrastive loss.
This loss drives each positive feature \(\mathbf{h}_{ic}\) to be closer to its own prototype than to others through a prototype-based softmax:
\begin{equation}
\mathcal{L}_{\text{Pos}}
= -\,\frac{1}{N}\sum_{i=1}^{N}
\frac{1}{|\mathcal{C}_i^{+}|}\sum_{c\in\mathcal{C}_i^{+}}
\log\frac{\exp\!\big(\mathbf{h}_{ic}^{\top}\mathbf{m}_{c}\big)}
{\sum_{r=1}^{C}\exp\!\big(\mathbf{h}_{ic}^{\top}\mathbf{m}_{r}\big)}.
\label{eq:pos}
\end{equation}

\noindent\textbf{Total Objective.}
The total training loss is defined as:
\begin{equation}
\mathcal{L}_{\text{total}}
\;=\;
\mathcal{L}_{\text{BCE}}
+\lambda_{1}\,\mathcal{L}_{\text{Neg}}
+\lambda_{2}\,\mathcal{L}_{\text{Pos}}.
\label{eq:total}
\end{equation}
\noindent where \(\lambda_{1},\lambda_{2}\ge 0\) are weighting coefficients that balance the contributions of the regularization terms.

\subsection{FedNCA-ML}
\label{sec:FedNCA-ML}
In summary, we propose FedNCA-ML for multi-label, label-skewed FL. 
Our design focuses on client-side training and introduces an NC-inspired pre-alignment strategy that leverages a shared geometric prior to anchor class-wise representations to consistent directions across clients, mitigating drift caused by heterogeneous local distributions. 
On the server, aggregation follows the standard FedAvg procedure, averaging client-updated weights in each communication round.
The overall pipeline is summarized in Algorithm~\ref{pseudo_code: algorithm_1}.
\begin{algorithm}[t]
\footnotesize
\setlength{\baselineskip}{0.8\baselineskip} % tighter vertical spacing
\caption{FedNCA-ML}
\textbf{Input:} $K$ clients with datasets $\{\mathcal{D}_k\}_{k=1}^K$; initial global model $w_0$; 
predefined ETF matrix $\mathbf{M}$; learning rate $\eta$; local epochs $E$; communication rounds $T$.
\begin{algorithmic}[1]
\State \textbf{Server executes:}
\State Initialize $w \gets w_0$
\For{$t = 0$ to $T-1$} \Comment{communication rounds}
    \For{each client $k \in \{1,\dots,K\}$ in parallel}
        \State $w_{t+1}^k \gets \textsc{ClientUpdate}(\mathcal{D}_k, w_t, \mathbf{M})$
    \EndFor
    \State $w_{t+1} \gets \frac{1}{K} \sum_{k=1}^{K} w_{t+1}^k$ \Comment{model aggregation}
    \State Broadcast $w_{t+1}$ to all clients
\EndFor
\Statex
\Function{ClientUpdate}{$\mathcal{D}_k, w, \mathbf{M}$}
    \For{$e = 0$ to $E-1$} \Comment{local epochs}
        \For{each batch $(x,y) \subset \mathcal{D}_k$}
            \State $\mathbf{F} \gets \textsc{FeatureExtractor}(w, x)$
            \State $\mathbf{H} \gets \textsc{LADM}(\mathbf{F}, \mathbf{M})$ \Comment{Eqs.~\ref{eq:LADM_F},~\ref{eq:LADM_reshape},~\ref{eq:LADM_position},~\ref{eq:LADM_ETF},~\ref{eq:LADM_h},~\ref{eq:LADM_H}}
            \State Compute prediction $\hat{y}_{c} = \sigma(\mathbf{h}_{c}^\top \mathbf{m}_c)$ \Comment{Eqs.~\ref{eq:sigmoid}}
            \State Compute loss $\mathcal{L}_{\text{total}}(w;\mathbf{M},\mathbf{H}, \mathbf{\hat{y}})$ \Comment{Eqs.\ref{eq:bce},\ref{eq:neg},\ref{eq:neg-loss},\ref{eq:pos},\ref{eq:total}}
            \State Update $w \gets w - \eta \nabla_w \mathcal{L}_{\text{total}}$
        \EndFor
    \EndFor
    \State \Return $w$
\EndFunction
\end{algorithmic}
\label{pseudo_code: algorithm_1}
\end{algorithm}
\vspace{-10pt}
\section{Experiments}
\label{sec: experiment}
% In this section, we perform experiments on different federated incremental scenarios on benchmark datasets. 
% Sufficient analyses are also performed to validate the effectiveness of each element of our proposed method.

\subsection{Dataset and Evaluation Metric}
\noindent\textbf{Datasets.}
We evaluate the proposed method on both general computer vision (CV) and medical imaging benchmarks.
For general CV, we use CIFAR-10~\cite{krizhevsky2009learning}, PASCAL VOC~\cite{everingham2010pascal}, and MS COCO~\cite{lin2014microsoft}.
Since CIFAR-10 is originally a multi-class dataset, we follow~\cite{li2023neural} to construct a multi-label variant by composing multiple images into a single composite image and using the union of their labels as the ground truth.
For medical imaging, we use DermaMNIST~\cite{yang2023medmnist} and ChestX-ray14~\cite{wang2017chestxray}.
DermaMNIST contains 7 skin disease categories in a single-class format and is converted to multi-label using the same strategy as CIFAR-10.
ChestX-ray14 is naturally multi-label, comprising 14 thoracic disease categories plus an additional ``No Finding'' label.
Since a significant portion of the dataset, 57\% of the training data, is ``No Finding'' samples (negative cases with all-zero labels), we distribute these samples evenly across all clients.
This setup mimics a realistic clinical scenario in which healthy cases are prevalent, while disease cases are relatively rare and unevenly distributed.
Detailed information about the datasets and local data distributions under various experimental settings is provided in the Appendix~\ref{sup_sec: dataset}.

\noindent\textbf{Evaluation Metric.}
Given our focus on label-skewed data distributions, we report both instance-wise (micro) and class-wise (macro) performance metrics.
The macro metric provides a balanced evaluation across classes, mitigating bias toward frequent categories.
Following standard practice, we report AUC and F1 scores for CIFAR-10, VOC, COCO, and DermaMNIST, and AUC for ChestX-ray14. 
Together, these metrics comprehensively evaluate overall performance and class-level behaviour.

\subsection{Task Setup and Implementation Details}
\textbf{Task Setup.}
We simulate an FL system with 10 clients to mimic a potential real-world clinical deployment and adopt full client participation in each round.
To model heterogeneity, we generate non-IID client distributions by partitioning data with a Dirichlet prior parameterized by the concentration factor $\beta$.
To further emulate label skew, we introduce a class-presence ratio $\gamma$, which restricts the set of classes observed by each client, simulating missing-class scenarios.

\noindent \textbf{Implementation Details.}
All experiments adopt ResNet-18~\cite{he2016deep} as the feature extractor.
We train each global model for 100 communication rounds with one local epoch per round, and select the final checkpoint based on the best validation performance.
We use a batch size of 32, an initial learning rate of $1\times10^{-4}$, and AdamW with weight decay 0.01.
Models on CIFAR-10 are trained from scratch, while models on the other datasets are initialized with ImageNet-pretrained weights.
Accordingly, we set the negative feature rejection coefficient $\lambda_1$ to 1 for CIFAR-10 to provide stronger regularisation during scratch training, and to 0.01 for pretrained models, which typically require milder regularisation.
The positive feature contrastive coefficient $\lambda_2$ is set to 1 across all experiments.
We repeat all experiments three times with different random seeds and report the mean and standard deviation.

\begin{table}[t]
    \caption{
        Comparisons on multi-label CIFAR-10~\cite{krizhevsky2009learning}.
        Missing-class scenarios are controlled by the class-presence ratio ($\gamma$), and non-IID client distributions are generated with a Dirichlet concentration parameter ($\beta$).
        We report both class-wise (macro) and instance-wise (micro) performance.
        }
    \centering
    \tiny
    \renewcommand{\arraystretch}{1.2} % Adjust row separation
    \resizebox{.99\linewidth}{!}{
    \begin{tabular}{l|cccc|cccc}
    \hline 

\multicolumn{1}{c|}{\multirow{2}{*}{\textbf{Method}}}
% \multirow{2}{*}{Method} 
& \multicolumn{4}{c|}{$\beta = 0.5$, $\gamma = 0.5$ ($\leq$ 5 of 10 classes/client)} 
& \multicolumn{4}{c}{$\beta = 0.1$, $\gamma = 0.5$ ($\leq$ 5 of 10 classes/client)} \\
\cline{2-9}
& macro-AUC & macro-F1 & micro-AUC & micro-F1 
& macro-AUC & macro-F1 & micro-AUC & micro-F1 \\
    \hline
Centralized & 92.20 \scalebox{0.75}{\textcolor{gray}{$\pm$0.36}} & 61.30 \scalebox{0.75}{\textcolor{gray}{$\pm$0.94}} & 90.04 \scalebox{0.75}{\textcolor{gray}{$\pm$0.65}} & 62.02 \scalebox{0.75}{\textcolor{gray}{$\pm$0.62}} & 92.20 \scalebox{0.75}{\textcolor{gray}{$\pm$0.36}} & 61.30 \scalebox{0.75}{\textcolor{gray}{$\pm$0.94}} & 90.04 \scalebox{0.75}{\textcolor{gray}{$\pm$0.65}} & 62.02 \scalebox{0.75}{\textcolor{gray}{$\pm$0.62}} \\
\hline
FedAvg \cite{mcmahan2017communication} & 82.29 \scalebox{0.75}{\textcolor{gray}{$\pm$0.46}}  & 39.47 \scalebox{0.75}{\textcolor{gray}{$\pm$1.59}} & 81.48 \scalebox{0.75}{\textcolor{gray}{$\pm$0.78}} & 40.26 \scalebox{0.75}{\textcolor{gray}{$\pm$1.09}} & 78.92 \scalebox{0.75}{\textcolor{gray}{$\pm$0.39}} & 31.17 \scalebox{0.75}{\textcolor{gray}{$\pm$0.51}} & 77.62 \scalebox{0.75}{\textcolor{gray}{$\pm$0.24}} & 35.07 \scalebox{0.75}{\textcolor{gray}{$\pm$0.47}} \\
FedCurv \cite{shoham2019overcoming} & 82.53 \scalebox{0.75}{\textcolor{gray}{$\pm$0.30}} & 39.96 \scalebox{0.75}{\textcolor{gray}{$\pm$1.28}} & 82.10 \scalebox{0.75}{\textcolor{gray}{$\pm$0.29}} & 40.34 \scalebox{0.75}{\textcolor{gray}{$\pm$1.04}} & 79.06 \scalebox{0.75}{\textcolor{gray}{$\pm$0.51}} & 31.00 \scalebox{0.75}{\textcolor{gray}{$\pm$1.21}} & 77.46 \scalebox{0.75}{\textcolor{gray}{$\pm$0.47}} & 35.03 \scalebox{0.75}{\textcolor{gray}{$\pm$0.54}} \\
FedProx \cite{li2020federated} & 82.45 \scalebox{0.75}{\textcolor{gray}{$\pm$0.34}} & 39.22 \scalebox{0.75}{\textcolor{gray}{$\pm$0.74}} & 81.77 \scalebox{0.75}{\textcolor{gray}{$\pm$0.48}} & 39.66 \scalebox{0.75}{\textcolor{gray}{$\pm$0.53}} & 78.82 \scalebox{0.75}{\textcolor{gray}{$\pm$0.55}} & 30.74 \scalebox{0.75}{\textcolor{gray}{$\pm$0.47}} & 77.40 \scalebox{0.75}{\textcolor{gray}{$\pm$0.31}} & 35.02 \scalebox{0.75}{\textcolor{gray}{$\pm$0.27}} \\
SCAFFOLD \cite{karimireddy2020scaffold} & 82.51 \scalebox{0.75}{\textcolor{gray}{$\pm$0.44}} & 39.98 \scalebox{0.75}{\textcolor{gray}{$\pm$1.41}} & 82.08 \scalebox{0.75}{\textcolor{gray}{$\pm$0.53}} & 40.26 \scalebox{0.75}{\textcolor{gray}{$\pm$1.26}} & 79.00 \scalebox{0.75}{\textcolor{gray}{$\pm$0.23}} & 31.38 \scalebox{0.75}{\textcolor{gray}{$\pm$0.31}} & 77.72 \scalebox{0.75}{\textcolor{gray}{$\pm$0.15}} & 35.54 \scalebox{0.75}{\textcolor{gray}{$\pm$0.16}} \\
SphereFed \cite{dong2022spherefed} & \underline{83.63} \scalebox{0.75}{\textcolor{gray}{$\pm$1.50}} & 42.58 \scalebox{0.75}{\textcolor{gray}{$\pm$3.20}} & \underline{83.50} \scalebox{0.75}{\textcolor{gray}{$\pm$1.87}} & 43.18 \scalebox{0.75}{\textcolor{gray}{$\pm$2.34}} & \underline{80.62} \scalebox{0.75}{\textcolor{gray}{$\pm$1.46}} & \underline{36.83} \scalebox{0.75}{\textcolor{gray}{$\pm$1.39}} & 78.37 \scalebox{0.75}{\textcolor{gray}{$\pm$0.95}} & 38.18 \scalebox{0.75}{\textcolor{gray}{$\pm$1.40}} \\
FedLGT \cite{liu2024language} & 83.52 \scalebox{0.75}{\textcolor{gray}{$\pm$0.68}} & \underline{43.60} \scalebox{0.75}{\textcolor{gray}{$\pm$1.68}} & 83.36 \scalebox{0.75}{\textcolor{gray}{$\pm$0.71}} & \underline{44.03} \scalebox{0.75}{\textcolor{gray}{$\pm$1.71}} & 80.54 \scalebox{0.75}{\textcolor{gray}{$\pm$0.43}} & 36.30 \scalebox{0.75}{\textcolor{gray}{$\pm$0.82}} & \textbf{80.65} \scalebox{0.75}{\textcolor{gray}{$\pm$0.68}} & \underline{39.24} \scalebox{0.75}{\textcolor{gray}{$\pm$0.71}} \\
\rowcolor{LightCyan} FedNCA-ML & \textbf{87.55} \scalebox{0.75}{\textcolor{gray}{$\pm$0.31}} & \textbf{48.17} \scalebox{0.75}{\textcolor{gray}{$\pm$1.65}} & \textbf{87.00} \scalebox{0.75}{\textcolor{gray}{$\pm$0.31}} & \textbf{48.61} \scalebox{0.75}{\textcolor{gray}{$\pm$1.64}} & \textbf{83.80} \scalebox{0.75}{\textcolor{gray}{$\pm$0.54}} & \textbf{38.09} \scalebox{0.75}{\textcolor{gray}{$\pm$0.62}} & \underline{78.90} \scalebox{0.75}{\textcolor{gray}{$\pm$0.66}} & \textbf{41.60} \scalebox{0.75}{\textcolor{gray}{$\pm$0.67}} \\
    \hline
    \end{tabular}}
    \label{tab: Exp_CIFAR10}

% ================================================================================================
\vspace{0.75em}
    \centering
    \caption{
    %Attent + ETF + 10 * Feat Contrast (PSC) + 0.01 Reject topK
    Comparisons on PASCAL VOC~\cite{everingham2010pascal}.
    %with the class presence ratio ($\gamma$) set to 0.5 ($\leq$ 10 of 20 classes per client).
    %This dataset contains 20 classes.
    }
    \renewcommand{\arraystretch}{1.2} % Adjust row separation
    
    \resizebox{.99\linewidth}{!}{
    \begin{tabular}{l|cccc|cccc}
    \hline 
\multicolumn{1}{c|}{\multirow{2}{*}{\textbf{Method}}}
    & \multicolumn{4}{c|}{$\beta = 0.05$, $\gamma = 0.5$ ($\leq$ 10 of 20 classes/client)} 
    & \multicolumn{4}{c}{$\beta = 0.01$, $\gamma = 0.5$ ($\leq$ 10 of 20 classes/client)} \\
    \cline{2-9}
    & macro-AUC & macro-F1 & micro-AUC & micro-F1 
    & macro-AUC & macro-F1 & micro-AUC & micro-F1 \\
    \hline
Centralized & 95.48 \scalebox{0.75}{\textcolor{gray}{$\pm$0.11}} & 74.61 \scalebox{0.75}{\textcolor{gray}{$\pm$0.10}} & 96.11 \scalebox{0.75}{\textcolor{gray}{$\pm$0.12}} & 76.94 \scalebox{0.75}{\textcolor{gray}{$\pm$0.09}} & 95.48 \scalebox{0.75}{\textcolor{gray}{$\pm$0.11}} & 74.61 \scalebox{0.75}{\textcolor{gray}{$\pm$0.10}} & 96.11 \scalebox{0.75}{\textcolor{gray}{$\pm$0.12}} & 76.94 \scalebox{0.75}{\textcolor{gray}{$\pm$0.09}} \\
\hline
FedAvg \cite{mcmahan2017communication} & 93.54 \scalebox{0.75}{\textcolor{gray}{$\pm$0.23}} & 57.57 \scalebox{0.75}{\textcolor{gray}{$\pm$0.76}} & \underline{94.13} \scalebox{0.75}{\textcolor{gray}{$\pm$0.33}} & 64.67 \scalebox{0.75}{\textcolor{gray}{$\pm$0.35}} & \underline{93.31} \scalebox{0.75}{\textcolor{gray}{$\pm$0.11}} & 47.73 \scalebox{0.75}{\textcolor{gray}{$\pm$1.03}} & 93.05 \scalebox{0.75}{\textcolor{gray}{$\pm$0.19}} & 60.46 \scalebox{0.75}{\textcolor{gray}{$\pm$0.43}} \\
FedCurv \cite{shoham2019overcoming} & 93.53 \scalebox{0.75}{\textcolor{gray}{$\pm$0.33}} & 57.36 \scalebox{0.75}{\textcolor{gray}{$\pm$0.41}} & 94.10 \scalebox{0.75}{\textcolor{gray}{$\pm$0.20}} & 64.60 \scalebox{0.75}{\textcolor{gray}{$\pm$0.26}} & 93.13 \scalebox{0.75}{\textcolor{gray}{$\pm$0.29}} & 48.54 \scalebox{0.75}{\textcolor{gray}{$\pm$0.68}} & \textbf{93.81} \scalebox{0.75}{\textcolor{gray}{$\pm$0.09}} & 60.49 \scalebox{0.75}{\textcolor{gray}{$\pm$1.16}} \\
FedProx \cite{li2020federated} & \underline{93.55} \scalebox{0.75}{\textcolor{gray}{$\pm$0.20}} & 57.04 \scalebox{0.75}{\textcolor{gray}{$\pm$0.24}} & 94.04 \scalebox{0.75}{\textcolor{gray}{$\pm$0.17}} & 64.43 \scalebox{0.75}{\textcolor{gray}{$\pm$0.29}} & 93.14 \scalebox{0.75}{\textcolor{gray}{$\pm$0.25}} & 49.49 \scalebox{0.75}{\textcolor{gray}{$\pm$0.68}} & \underline{93.80} \scalebox{0.75}{\textcolor{gray}{$\pm$0.17}} & 62.18 \scalebox{0.75}{\textcolor{gray}{$\pm$0.51}} \\
SCAFFOLD \cite{karimireddy2020scaffold} & 93.17 \scalebox{0.75}{\textcolor{gray}{$\pm$0.20}} & 57.44 \scalebox{0.75}{\textcolor{gray}{$\pm$0.33}} & 94.03 \scalebox{0.75}{\textcolor{gray}{$\pm$0.11}} & 64.95 \scalebox{0.75}{\textcolor{gray}{$\pm$0.21}} & \textbf{93.41} \scalebox{0.75}{\textcolor{gray}{$\pm$0.17}} & 49.64 \scalebox{0.75}{\textcolor{gray}{$\pm$0.77}} & 93.60 \scalebox{0.75}{\textcolor{gray}{$\pm$0.21}} & 61.44 \scalebox{0.75}{\textcolor{gray}{$\pm$0.56}} \\
SphereFed \cite{dong2022spherefed} & 83.72 \scalebox{0.75}{\textcolor{gray}{$\pm$1.82}} & 33.49 \scalebox{0.75}{\textcolor{gray}{$\pm$1.15}} & 84.38 \scalebox{0.75}{\textcolor{gray}{$\pm$1.10}} & 38.19 \scalebox{0.75}{\textcolor{gray}{$\pm$1.17}} & 84.25 \scalebox{0.75}{\textcolor{gray}{$\pm$2.44}} & 32.44 \scalebox{0.75}{\textcolor{gray}{$\pm$0.21}} & 85.94 \scalebox{0.75}{\textcolor{gray}{$\pm$3.67}} & 35.18 \scalebox{0.75}{\textcolor{gray}{$\pm$1.30}} \\
FedLGT \cite{liu2024language} & 91.93 \scalebox{0.75}{\textcolor{gray}{$\pm$0.42}} & \underline{62.11} \scalebox{0.75}{\textcolor{gray}{$\pm$0.44}} & 91.75 \scalebox{0.75}{\textcolor{gray}{$\pm$0.42}} & \underline{67.58} \scalebox{0.75}{\textcolor{gray}{$\pm$0.45}} & 91.25 \scalebox{0.75}{\textcolor{gray}{$\pm$0.27}} & \underline{56.53} \scalebox{0.75}{\textcolor{gray}{$\pm$2.42}} & 91.46 \scalebox{0.75}{\textcolor{gray}{$\pm$0.41}} & \underline{63.51} \scalebox{0.75}{\textcolor{gray}{$\pm$1.73}} \\
\rowcolor{LightCyan} FedNCA-ML & \textbf{93.82} \scalebox{0.75}{\textcolor{gray}{$\pm$0.36}} & \textbf{64.28} \scalebox{0.75}{\textcolor{gray}{$\pm$0.05}} & \textbf{94.52} \scalebox{0.75}{\textcolor{gray}{$\pm$0.15}} & \textbf{67.61} \scalebox{0.75}{\textcolor{gray}{$\pm$0.49}} & 93.01 \scalebox{0.75}{\textcolor{gray}{$\pm$0.22}} & \textbf{61.08} \scalebox{0.75}{\textcolor{gray}{$\pm$0.10}} & 93.70 \scalebox{0.75}{\textcolor{gray}{$\pm$0.18}} & \textbf{65.05} \scalebox{0.75}{\textcolor{gray}{$\pm$0.53}} \\
    \hline
    \end{tabular}}
    \label{tab: Exp_VOC}
    \vspace{-3mm}
\end{table}

\begin{table}[t]
    \caption{
        Comparisons on MS COCO~\cite{lin2014microsoft}.
        }
    \setlength{\tabcolsep}{3pt} % Adjust column separation
    \centering
    \tiny
    \resizebox{.59\linewidth}{!}{
    \renewcommand{\arraystretch}{1.2} % Adjust row separation
    \begin{tabular}{l|cccc}
    \hline 

\multicolumn{1}{c|}{\multirow{2}{*}{\textbf{Method}}}
% \multirow{2}{*}{Method} 
& \multicolumn{4}{c}{$\beta = 0.05$, $\gamma = 0.75$ ($\leq$ 60 of 80 classes/client)} \\
\cline{2-5}
& macro-AUC & macro-F1 & micro-AUC & micro-F1 \\
    \hline
Centralized & 94.29 \scalebox{0.75}{\textcolor{gray}{$\pm$0.13}} & 58.43 \scalebox{0.75}{\textcolor{gray}{$\pm$0.02}} & 95.60 \scalebox{0.75}{\textcolor{gray}{$\pm$0.20}} & 64.40 \scalebox{0.75}{\textcolor{gray}{$\pm$0.50}}  \\
\hline
FedAvg \cite{mcmahan2017communication} & 93.66 \scalebox{0.75}{\textcolor{gray}{$\pm$0.16}}  & 53.21 \scalebox{0.75}{\textcolor{gray}{$\pm$0.49}} & \underline{94.74} \scalebox{0.75}{\textcolor{gray}{$\pm$0.12}} & 60.35 \scalebox{0.75}{\textcolor{gray}{$\pm$0.15}} \\
FedCurv \cite{shoham2019overcoming} & \underline{94.03} \scalebox{0.75}{\textcolor{gray}{$\pm$0.11}} & 54.34 \scalebox{0.75}{\textcolor{gray}{$\pm$0.52}} & 95.40 \scalebox{0.75}{\textcolor{gray}{$\pm$0.32}} & 60.82 \scalebox{0.75}{\textcolor{gray}{$\pm$0.21}} \\
FedProx \cite{li2020federated} & 93.74 \scalebox{0.75}{\textcolor{gray}{$\pm$0.15}} & 53.65 \scalebox{0.75}{\textcolor{gray}{$\pm$0.27}} & \textbf{94.87} \scalebox{0.75}{\textcolor{gray}{$\pm$0.10}} & 60.58 \scalebox{0.75}{\textcolor{gray}{$\pm$0.15}} \\
SCAFFOLD \cite{karimireddy2020scaffold} & 93.55 \scalebox{0.75}{\textcolor{gray}{$\pm$0.22}} & 53.50 \scalebox{0.75}{\textcolor{gray}{$\pm$0.82}} & 94.68 \scalebox{0.75}{\textcolor{gray}{$\pm$0.12}} & 60.15 \scalebox{0.75}{\textcolor{gray}{$\pm$0.86}} \\
SphereFed \cite{dong2022spherefed} & 84.87 \scalebox{0.75}{\textcolor{gray}{$\pm$1.13}} & 35.26 \scalebox{0.75}{\textcolor{gray}{$\pm$0.48}} & 80.62 \scalebox{0.75}{\textcolor{gray}{$\pm$0.56}} & 50.58 \scalebox{0.75}{\textcolor{gray}{$\pm$0.72}} \\
FedLGT \cite{liu2024language} & 90.90 \scalebox{0.75}{\textcolor{gray}{$\pm$0.25}} & \underline{55.68} \scalebox{0.75}{\textcolor{gray}{$\pm$0.83}} & 92.37 \scalebox{0.75}{\textcolor{gray}{$\pm$0.32}} & \textbf{62.76} \scalebox{0.75}{\textcolor{gray}{$\pm$0.76}} \\
\rowcolor{LightCyan} FedNCA-ML & \textbf{94.10} \scalebox{0.75}{\textcolor{gray}{$\pm$0.18}} & \textbf{56.28} \scalebox{0.75}{\textcolor{gray}{$\pm$0.32}} & 94.71 \scalebox{0.75}{\textcolor{gray}{$\pm$0.20}} & \underline{61.71} \scalebox{0.75}{\textcolor{gray}{$\pm$0.52}} \\
\hline
\end{tabular}}
\label{tab: Exp_COCO}
\end{table}
\begin{table}[t]
    \centering
    \caption{
        Comparisons on multi-label DermaMNIST~\cite{yang2023medmnist}.
        %with the class presence ratio ($\gamma$) set to 0.71 ($\leq$ 5 of 7 classes per client).
        %Comparisons on the multi-label DermaMNIST dataset \cite{yang2023medmnist}. 
        % This dataset contains 7 classes of different skin types.
    }
    \renewcommand{\arraystretch}{1.2} % Adjust row separation
    \tiny
    \resizebox{.99\linewidth}{!}{
   \begin{tabular}{l|cccc|cccc}
    \hline 
\multicolumn{1}{c|}{\multirow{2}{*}{\textbf{Method}}}
& \multicolumn{4}{c|}{$\beta = 0.5$, $\gamma = 0.71$ ($\leq$ 5 of 7 classes/client)} 
& \multicolumn{4}{c}{$\beta = 0.1$, $\gamma = 0.71$ ($\leq$ 5 of 7 classes/client)} \\
\cline{2-9}
& macro-AUC & macro-F1 & micro-AUC & micro-F1 
& macro-AUC & macro-F1 & micro-AUC & micro-F1 \\
    \hline
Centralized & 91.83 \scalebox{0.75}{\textcolor{gray}{$\pm$0.40}} & 64.38 \scalebox{0.75}{\textcolor{gray}{$\pm$1.01}} & 94.48 \scalebox{0.75}{\textcolor{gray}{$\pm$0.25}} & 74.12 \scalebox{0.75}{\textcolor{gray}{$\pm$0.70}} & 91.83 \scalebox{0.75}{\textcolor{gray}{$\pm$0.40}} & 64.38 \scalebox{0.75}{\textcolor{gray}{$\pm$1.01}} & 94.48 \scalebox{0.75}{\textcolor{gray}{$\pm$0.25}} & 74.12 \scalebox{0.75}{\textcolor{gray}{$\pm$0.70}} \\
\hline
FedAvg \cite{mcmahan2017communication} & \underline{89.72} \scalebox{0.75}{\textcolor{gray}{$\pm$0.29}} & 54.88 \scalebox{0.75}{\textcolor{gray}{$\pm$0.97}} & \underline{92.51} \scalebox{0.75}{\textcolor{gray}{$\pm$0.36}} & \textbf{68.29} \scalebox{0.75}{\textcolor{gray}{$\pm$0.45}} & 83.88 \scalebox{0.75}{\textcolor{gray}{$\pm$1.21}} & 42.79 \scalebox{0.75}{\textcolor{gray}{$\pm$2.03}} & 87.23 \scalebox{0.75}{\textcolor{gray}{$\pm$0.86}} & 62.81 \scalebox{0.75}{\textcolor{gray}{$\pm$0.83}} \\
FedCurv \cite{shoham2019overcoming} & 89.48 \scalebox{0.75}{\textcolor{gray}{$\pm$0.18}} & 54.92 \scalebox{0.75}{\textcolor{gray}{$\pm$1.11}} & 92.26 \scalebox{0.75}{\textcolor{gray}{$\pm$0.36}} & 67.97 \scalebox{0.75}{\textcolor{gray}{$\pm$0.92}} & \underline{85.53} \scalebox{0.75}{\textcolor{gray}{$\pm$1.37}} & 43.45 \scalebox{0.75}{\textcolor{gray}{$\pm$1.20}} & 88.87 \scalebox{0.75}{\textcolor{gray}{$\pm$0.50}} & \textbf{64.37} \scalebox{0.75}{\textcolor{gray}{$\pm$0.26}} \\
FedProx \cite{li2020federated} & 89.69 \scalebox{0.75}{\textcolor{gray}{$\pm$0.31}} & 55.78 \scalebox{0.75}{\textcolor{gray}{$\pm$0.94}} & 92.06 \scalebox{0.75}{\textcolor{gray}{$\pm$0.46}} & 68.01 \scalebox{0.75}{\textcolor{gray}{$\pm$1.46}} & 84.45 \scalebox{0.75}{\textcolor{gray}{$\pm$1.34}} & 43.54 \scalebox{0.75}{\textcolor{gray}{$\pm$1.35}} & 88.28 \scalebox{0.75}{\textcolor{gray}{$\pm$1.07}} & 63.25 \scalebox{0.75}{\textcolor{gray}{$\pm$0.63}} \\
SCAFFOLD \cite{karimireddy2020scaffold} & \underline{89.72} \scalebox{0.75}{\textcolor{gray}{$\pm$0.49}} & \underline{55.85} \scalebox{0.75}{\textcolor{gray}{$\pm$0.68}} & 92.45 \scalebox{0.75}{\textcolor{gray}{$\pm$0.32}} & 68.09 \scalebox{0.75}{\textcolor{gray}{$\pm$0.59}} & 84.59 \scalebox{0.75}{\textcolor{gray}{$\pm$1.03}} & 43.20 \scalebox{0.75}{\textcolor{gray}{$\pm$1.00}} & 88.91 \scalebox{0.75}{\textcolor{gray}{$\pm$0.90}} & \underline{63.82} \scalebox{0.75}{\textcolor{gray}{$\pm$1.16}} \\
SphereFed \cite{dong2022spherefed} & 85.09 \scalebox{0.75}{\textcolor{gray}{$\pm$0.87}} & 43.41 \scalebox{0.75}{\textcolor{gray}{$\pm$1.73}} & 89.65 \scalebox{0.75}{\textcolor{gray}{$\pm$0.55}} & 65.24 \scalebox{0.75}{\textcolor{gray}{$\pm$0.58}} & 81.78 \scalebox{0.75}{\textcolor{gray}{$\pm$1.14}} & 40.90 \scalebox{0.75}{\textcolor{gray}{$\pm$1.79}} & 86.59 \scalebox{0.75}{\textcolor{gray}{$\pm$1.87}} & 54.94 \scalebox{0.75}{\textcolor{gray}{$\pm$1.12}} \\
FedLGT \cite{liu2024language} & 87.63 \scalebox{0.75}{\textcolor{gray}{$\pm$0.71}} & 55.82 \scalebox{0.75}{\textcolor{gray}{$\pm$1.34}} & 91.19 \scalebox{0.75}{\textcolor{gray}{$\pm$0.51}} & 67.93 \scalebox{0.75}{\textcolor{gray}{$\pm$0.45}} & 84.91 \scalebox{0.75}{\textcolor{gray}{$\pm$0.79}} & \underline{45.61} \scalebox{0.75}{\textcolor{gray}{$\pm$1.16}} & \underline{89.42} \scalebox{0.75}{\textcolor{gray}{$\pm$0.65}} & 61.15 \scalebox{0.75}{\textcolor{gray}{$\pm$2.43}} \\
\rowcolor{LightCyan} FedNCA-ML & \textbf{90.12} \scalebox{0.75}{\textcolor{gray}{$\pm$0.51}} & \textbf{56.31} \scalebox{0.75}{\textcolor{gray}{$\pm$0.65}} & \textbf{92.85} \scalebox{0.75}{\textcolor{gray}{$\pm$0.48}} & \underline{68.20} \scalebox{0.75}{\textcolor{gray}{$\pm$0.38}} & \textbf{86.30} \scalebox{0.75}{\textcolor{gray}{$\pm$0.85}} & \textbf{50.54} \scalebox{0.75}{\textcolor{gray}{$\pm$1.37}} & \textbf{89.74} \scalebox{0.75}{\textcolor{gray}{$\pm$0.98}} & 63.76 \scalebox{0.75}{\textcolor{gray}{$\pm$1.31}} \\
    \hline
    \end{tabular}}
    \label{tab: Exp_DermaMNIST}
    \vspace{-3mm}
\end{table}

\begin{table}[t]
    \caption{
    Comparisons on ChestX-ray14~\cite{wang2017chestxray} with $\leq$ 7 of 14 disease classes/client.
    % This dataset contains 14 related chest diseases for the chest X-ray scan. 
    % Healthy data equally distributed.
    }
    \centering
    \tiny
    \renewcommand{\arraystretch}{1.2} % Adjust row separation
    \resizebox{.60\linewidth}{!}{
    \begin{tabular}{l | c c | c c}
    \hline 
    
\multicolumn{1}{c|}{\multirow{2}{*}{\textbf{Method}}}
    & \multicolumn{2}{c|}{$\beta = 0.5$, $\gamma = 0.5$} & \multicolumn{2}{c}{$\beta = 0.1$,  $\gamma = 0.5$} \\
    \cline{2-5}
    & macro-AUC & micro-AUC & macro-AUC & micro-AUC \\
    \hline
Centralized & 71.66 \scalebox{0.75}{\textcolor{gray}{$\pm$0.12}} & 79.54 \scalebox{0.75}{\textcolor{gray}{$\pm$0.36}} & 71.66 \scalebox{0.75}{\textcolor{gray}{$\pm$0.12}} & 79.54 \scalebox{0.75}{\textcolor{gray}{$\pm$0.36}} \\
\hline
FedAvg \cite{mcmahan2017communication} & 69.02 \scalebox{0.75}{\textcolor{gray}{$\pm$0.24}} & \textbf{73.18} \scalebox{0.75}{\textcolor{gray}{$\pm$0.15}} & 69.05 \scalebox{0.75}{\textcolor{gray}{$\pm$0.78}} & \textbf{77.90} \scalebox{0.75}{\textcolor{gray}{$\pm$0.20}} \\
FedCurv \cite{shoham2019overcoming} & 68.72 \scalebox{0.75}{\textcolor{gray}{$\pm$1.19}} & 72.15 \scalebox{0.75}{\textcolor{gray}{$\pm$0.17}} & 69.34 \scalebox{0.75}{\textcolor{gray}{$\pm$0.48}} & 77.36 \scalebox{0.75}{\textcolor{gray}{$\pm$0.27}} \\
FedProx \cite{li2020federated} & 69.02 \scalebox{0.75}{\textcolor{gray}{$\pm$0.21}} & 72.04 \scalebox{0.75}{\textcolor{gray}{$\pm$0.59}} & 69.59 \scalebox{0.75}{\textcolor{gray}{$\pm$0.23}} & 77.43 \scalebox{0.75}{\textcolor{gray}{$\pm$0.24}} \\
SCAFFOLD \cite{karimireddy2020scaffold}  & 69.42 \scalebox{0.75}{\textcolor{gray}{$\pm$0.39}} & \underline{72.49} \scalebox{0.75}{\textcolor{gray}{$\pm$0.38}} & 67.45 \scalebox{0.75}{\textcolor{gray}{$\pm$0.54}} & \textbf{77.90} \scalebox{0.75}{\textcolor{gray}{$\pm$0.69}} \\
SphereFed \cite{dong2022spherefed} & 58.35 \scalebox{0.75}{\textcolor{gray}{$\pm$0.57}} & 69.51 \scalebox{0.75}{\textcolor{gray}{$\pm$0.56}} & 61.96 \scalebox{0.75}{\textcolor{gray}{$\pm$0.16}} & 73.59 \scalebox{0.75}{\textcolor{gray}{$\pm$1.02}} \\
FedLGT \cite{liu2024language} & \underline{69.86} \scalebox{0.75}{\textcolor{gray}{$\pm$0.76}} & 72.27 \scalebox{0.75}{\textcolor{gray}{$\pm$1.07}} & \underline{70.16} \scalebox{0.75}{\textcolor{gray}{$\pm$0.37}} & 77.67 \scalebox{0.75}{\textcolor{gray}{$\pm$0.57}} \\
\rowcolor{LightCyan} FedNCA-ML & \textbf{70.55} \scalebox{0.75}{\textcolor{gray}{$\pm$0.15}} & 71.28 \scalebox{0.75}{\textcolor{gray}{$\pm$1.34}} & \textbf{71.28} \scalebox{0.75}{\textcolor{gray}{$\pm$0.15}} & \underline{77.86} \scalebox{0.75}{\textcolor{gray}{$\pm$0.45}} \\
\hline
    \end{tabular}}
    \label{tab: Exp_Chest}
\end{table}

\subsection{Performance Comparison}
To evaluate the effectiveness of the proposed method, we compare FedNCA-ML with state-of-the-art approaches on five datasets under nine label-skewed FL settings.
As summarized in Tables~\ref{tab: Exp_CIFAR10}, \ref{tab: Exp_VOC}, \ref{tab: Exp_COCO}, \ref{tab: Exp_DermaMNIST}, and \ref{tab: Exp_Chest}, FedNCA-ML achieves the best class-wise performance in most cases, highlighting its ability to deliver balanced and generalizable predictions under heterogeneous FL distributions.
Specifically, on multi-label CIFAR-10 (Tables~\ref{tab: Exp_CIFAR10}), under non-IID Dirichlet settings of $\beta=0.5$ ($\beta=0.1$) with a maximum of 5 out of 10 classes per client, FedNCA-ML surpasses the second-best approach by 3.92\% (3.18\%) in class-wise AUC and 4.57\% (1.26\%) in class-wise F1 score.
On multi-label DermaMNIST Tables~\ref{tab: Exp_DermaMNIST}, under $\beta=0.5$ ($\beta=0.1$) with up to 5 of 7 classes per client, it achieves improvements of 0.40\% (0.77\%) in AUC and 0.46\% (4.93\%) in F1 score.
On VOC (Tables~\ref{tab: Exp_VOC}), under more challenging settings of $\beta=0.05$ ($\beta=0.01$) with up to 10 of 20 classes per client, FedNCA-ML outperforms the second-best approach by 2.17\% (4.55\%) in class-wise F1 score.
On COCO (Tables~\ref{tab: Exp_COCO}), under $\beta=0.05$ with up to 60 of 80 classes per client, it yields 0.60\% improvement in class-wise F1 score.

\begin{figure}[t]
    \centering
    \begin{subfigure}[t]{0.32\linewidth}
        \includegraphics[width=\linewidth, trim={1.5cm 2cm 6.2cm 0.6cm}, clip]{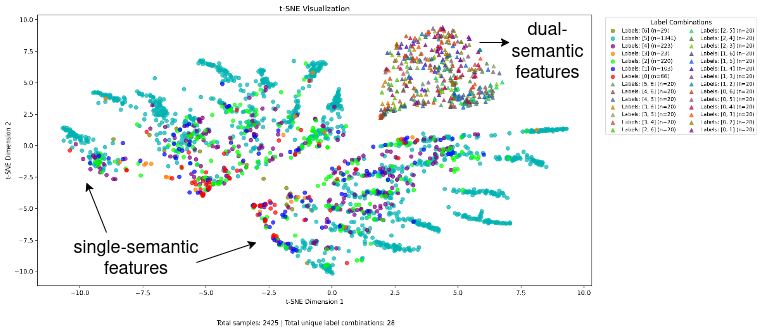}  % left bottom right top
        \subcaption{\tiny Learnable FC classifier}
        \label{fig: Ablation_Derma-MNIST_tsne_a}
    \end{subfigure}
    \begin{subfigure}[t]{0.32\linewidth}
        \includegraphics[width=\linewidth, trim={2.5cm 3cm 10cm 1cm}, clip]{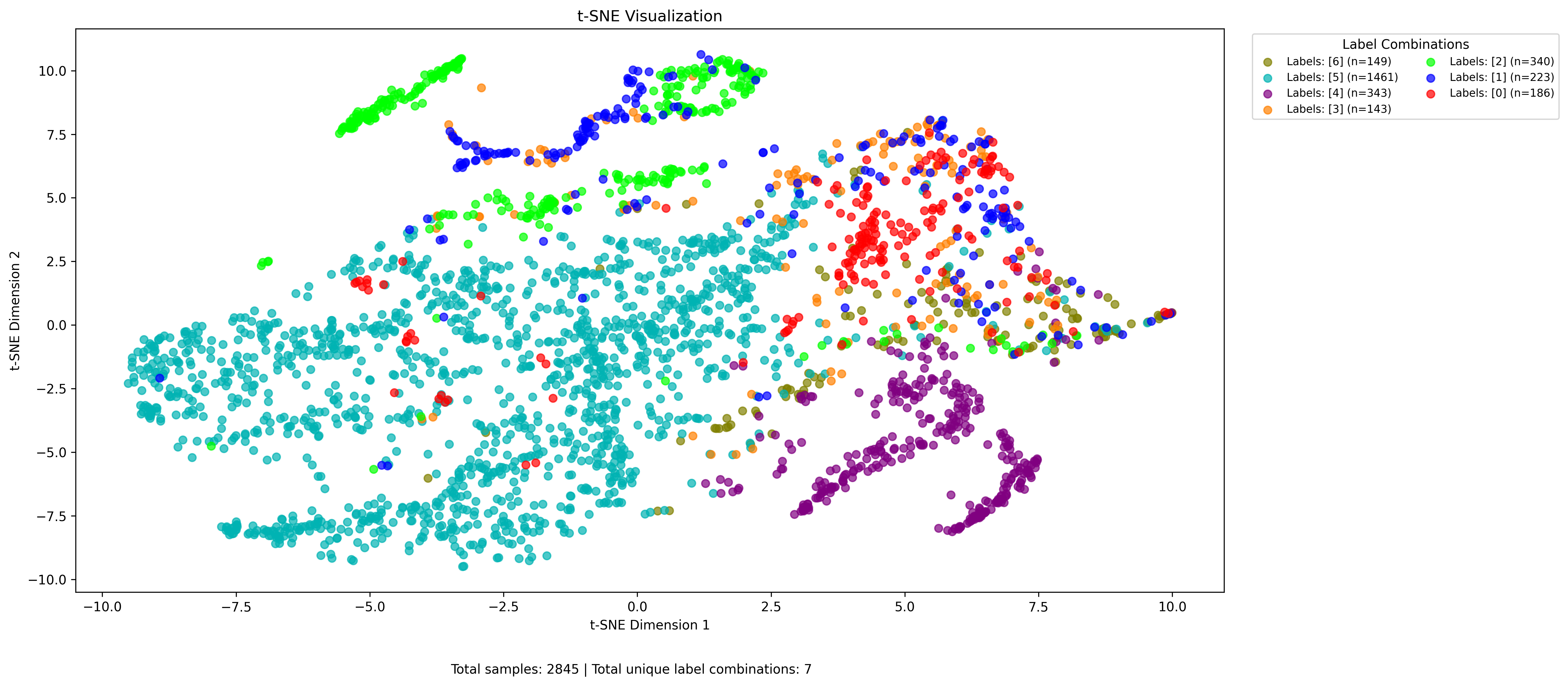}
        \subcaption{\tiny LADM $+$ fixed ETF classifier}
        \label{fig: Ablation_Derma-MNIST_tsne_b}
    \end{subfigure}
    \begin{subfigure}[t]{0.32\linewidth}
        \includegraphics[width=\linewidth, trim={2.5cm 3cm 10cm 1cm}, clip]{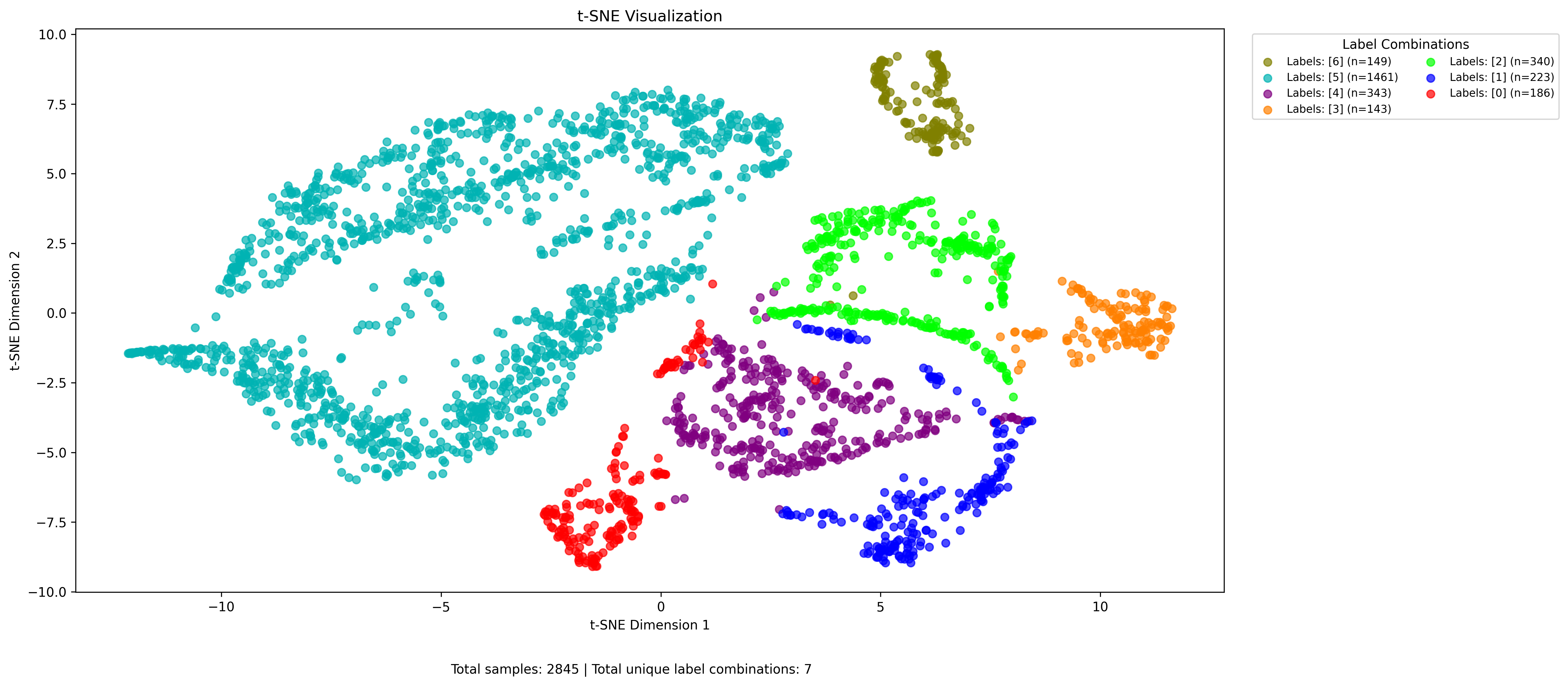}
        \subcaption{\tiny LADM $+$ fixed ETF $+$ Dual regularization}
        \label{fig: Ablation_Derma-MNIST_tsne_c}
    \end{subfigure}
    \caption{
    t-SNE visualisation of test data feature embeddings on the multi-label DermaMNIST experiment with $\beta = 0.1$ and $\gamma = 0.71$.
    Each colour represents a class.
    Observing from subfigure (a), without class-wise feature extraction (LADM), the model relies on undesired information, such as the number of labels per sample, for clustering.
    }
    \label{fig: Ablation_Derma-MNIST_tsne}
\end{figure}

%\definecolor{blue1}{RGB}{235,245,255}
%\definecolor{blue2}{RGB}{225,240,255}
%\definecolor{blue3}{RGB}{215,235,255}
%\definecolor{blue4}{RGB}{205,230,255}
%\definecolor{blue5}{RGB}{195,225,255}
%\definecolor{blue6}{RGB}{185,220,255}
%\definecolor{blue7}{RGB}{175,215,255}

\definecolor{blue1}{RGB}{235,245,255}  % 144
\definecolor{blue2}{RGB}{232,243,255}  % 144+149
\definecolor{blue3}{RGB}{228,241,255}  % +186
\definecolor{blue4}{RGB}{223,239,255}  % +223
\definecolor{blue5}{RGB}{215,235,255}  % +340
\definecolor{blue6}{RGB}{207,231,255}  % +343
\definecolor{blue7}{RGB}{175,215,255}  % +1461 (largest jump)

\begin{table}[t]
    \centering
    \tiny
    \caption{
    Ablation study of the proposed method on multi-label DermaMNIST with $\beta=0.1$ and $\gamma=0.71$.
    We further report the F1 score for each class.
    The blue shading indicates the class prevalence, ranging from the majority class (60.25\%) to the rarest class (5.90\%).
    Dataset details are provided in Appendix~\ref{sup_sec: dataset}.
    }
    \renewcommand{\arraystretch}{1.2} % Adjust row separation
    \resizebox{.99\linewidth}{!}{
    \begin{tabular}{c c c c | c c c c |
    >{\columncolor{blue3}}c
    >{\columncolor{blue4}}c
    >{\columncolor{blue5}}c
    >{\columncolor{blue1}}c
    >{\columncolor{blue6}}c
    >{\columncolor{blue7}}c
    >{\columncolor{blue2}}c}
    \hline 
    ETF Clf & LADM & $\mathcal{L}_{\text{Neg}}$ & $\mathcal{L}_{\text{Pos}}$ & macro-AUC & macro-F1 & micro-AUC & micro-F1 & \multicolumn{7}{c}{F1 score for each class} \\
    \hline
    & & & & 83.95 & 40.69 & 86.68 & 63.03 & 44.51 & 10.28 & \textbf{34.88} & 52.23 & 1.15 & 82.28 & 60.63 \\
    \checkmark & & & & 83.61 & 33.35 & 87.26 & 61.48 & 28.09 & 15.87 & 2.31 & 48.00 & 0.00 & 81.36 & 57.80 \\
    \checkmark & \checkmark & & & 84.38 & 47.95 & 86.70 & 60.49 & \textbf{48.16} & 26.11 & 23.83 & 49.06 & 30.27 & 78.88 & \textbf{79.32} \\
    \checkmark & \checkmark & \checkmark & & 84.73 & 45.06 & 89.71 & 62.89 & 48.11 & 6.72 & 20.81 & 50.16 & 32.14 & 82.54 & 74.90 \\
    \checkmark & \checkmark & & \checkmark & 85.20 & 49.84 & 89.03 & 61.06 & 46.70 & \textbf{39.83} & 33.57 & 49.38 & 28.51 & 79.77 & 71.11 \\
    \checkmark & \checkmark & \checkmark & \checkmark & \textbf{87.69} & \textbf{51.38} & \textbf{90.36} & \textbf{63.26} & 45.65 & 35.35 & 28.41 & \textbf{53.45} & \textbf{36.01} & \textbf{83.00} & 77.82 \\
    \hline
    \end{tabular}}
    \label{tab: Ablation_Derma-MNIST}

% ================================================================================================
\vspace{1em}
    \centering
    \scriptsize
    \caption{
    Ablation study of the class-wise feature extraction block (LADM) on the multi-label DermaMNIST dataset with $\beta = 0.1$ and $\gamma = 0.71$.
    }
    \resizebox{.65\linewidth}{!}{
    \renewcommand{\arraystretch}{1.1} % Adjust row separation
    \begin{tabular}{c c | c c c c }
    \hline 
    query type & query init & macro-AUC & macro-F1 & micro-AUC & micro-F1 \\
    \hline
    learnable & random & 82.94 & 47.94 & 86.33 & 57.37 \\
    learnable & ETF & \textbf{85.39} & 46.92 & 84.94 & 53.37 \\
    fixed & ETF & 84.38 & \textbf{47.95} & \textbf{86.70} & \textbf{60.49} \\
    \hline
    \end{tabular}}
    \label{tab: Ablation_LADM_Derma-MNIST}
\end{table}

On ChestX-ray14 (Tables~\ref{tab: Exp_Chest}), under non-IID Dirichlet settings of $\beta = 0.5$ and $\beta = 0.1$ with up to 7 of 14 disease classes per client, FedNCA-ML improves class-wise AUC by 0.69\% and 1.21\%, respectively.
However, it achieves slightly lower overall AUC than some other methods.
ChestX-ray14 is highly imbalanced with 57\% of training and 38\% of testing samples labeled as ``No Finding''.
This severe imbalance leads many methods to overpredict the majority class, as reflected in a large gap between class-wise and overall AUC, indicating bias toward majority classes and degraded performance on minority (disease) classes.
In medical diagnosis, false positives are generally more tolerable than false negatives, especially for rare diseases.
The higher class-wise AUC and the smaller gap between overall and class-wise AUC achieved by FedNCA-ML suggest more balanced predictions and better recognition of minority disease classes.

\subsection{Ablation Study}
\label{sec: ablation_study}
\noindent\textbf{FedNCA-ML Component Analysis.}
We conduct an ablation study on the multi-label DermaMNIST dataset, which exhibits severe class imbalance (the majority class ``melanoma'' accounts for 61.97\% of the training set and 60.25\% of the test set) and pronounced inter-/intra-class variability due to diverse dermatological conditions.
We simulate heterogeneous clients with $\beta=0.1$ and $\gamma=0.71$, which further amplifies class imbalance and label-relation heterogeneity.
Results are reported in Table~\ref{tab: Ablation_Derma-MNIST}.
As shown in the table, using only the predefined ETF classifier degrades performance, because image-level features entangle multiple semantic cues and directly clustering such representations can misguide optimization.
In contrast, adding LADM for class-specific feature extraction and applying ETF anchoring to these class-specific features improves class-wise AUC by 0.43\% and class-wise F1 by 7.26\%.
Notably, the lowest per-class F1 increases from 1.15\% to 30.27\% with an improvement of 29.12\%.
Finally, introducing the regularisation terms further strengthens discriminative learning and yields more balanced class-wise performance, bringing additional gains of 3.31\% in class-wise AUC and 3.43\% in class-wise F1.
\begin{wrapfigure}{l}{0.6\linewidth}
    \centering
    \begin{subfigure}[t]{0.305\linewidth}
        \includegraphics[width=\linewidth, trim={0.2cm 0cm 4.3cm 0.785cm}, clip]{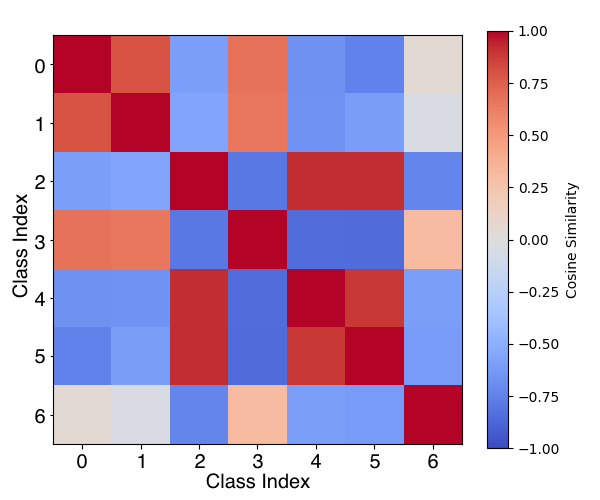}
        \subcaption{\tiny FC classifier}
        \label{fig: Ablation_Derma-MNIST_prototype_a}
    \end{subfigure}
    \begin{subfigure}[t]{0.305\linewidth}
        \includegraphics[width=\linewidth, trim={0.2cm 0cm 4.3cm 0.785cm}, clip]{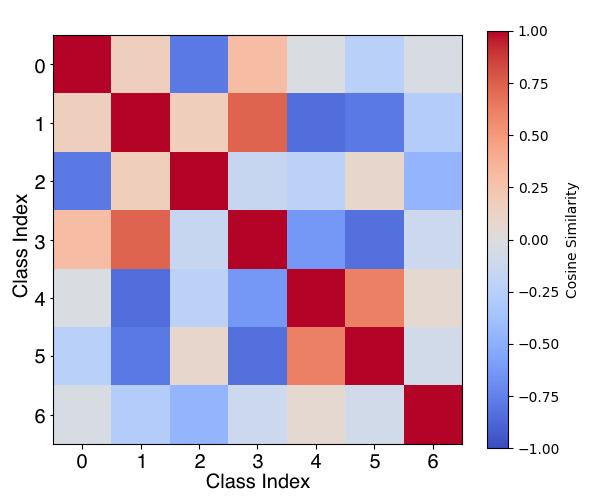}
        \subcaption{\tiny LADM + fixed ETF classifier}
        \label{fig: Ablation_Derma-MNIST_prototype_b}
    \end{subfigure}
    \begin{subfigure}[t]{0.36\linewidth}
        \includegraphics[width=\linewidth, trim={0.2cm 0cm 1.2cm 0.785cm}, clip]{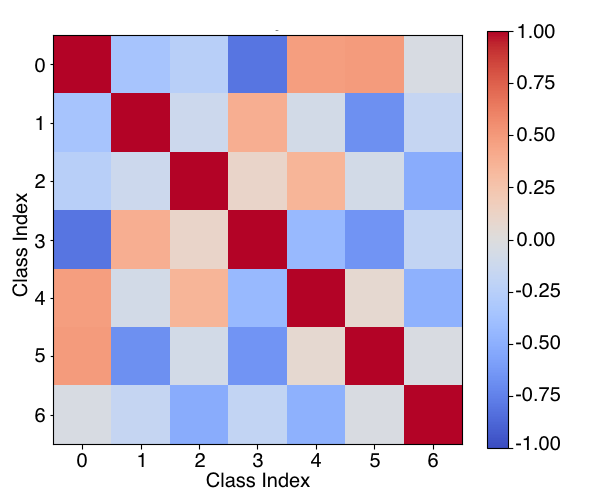}
        \subcaption{\tiny LADM + ETF + Dual regularization}
        \label{fig: Ablation_Derma-MNIST_prototype_c}
    \end{subfigure}
    \caption{
    Pairwise cosine similarity of class-wise average feature prototypes. 
    Incorporating LADM, ETF-based alignment, and structure-preserving regularization lowers inter-class similarity, reflecting enhanced separability and discrimination.
    }
    \label{fig: Ablation_Derma-MNIST_prototype}
    \vspace{-6mm}
\end{wrapfigure}

\begin{figure}[!h]
    \centering
    \includegraphics[width=0.9\linewidth, trim={0cm 0.2cm 0cm 0.1cm}, clip]{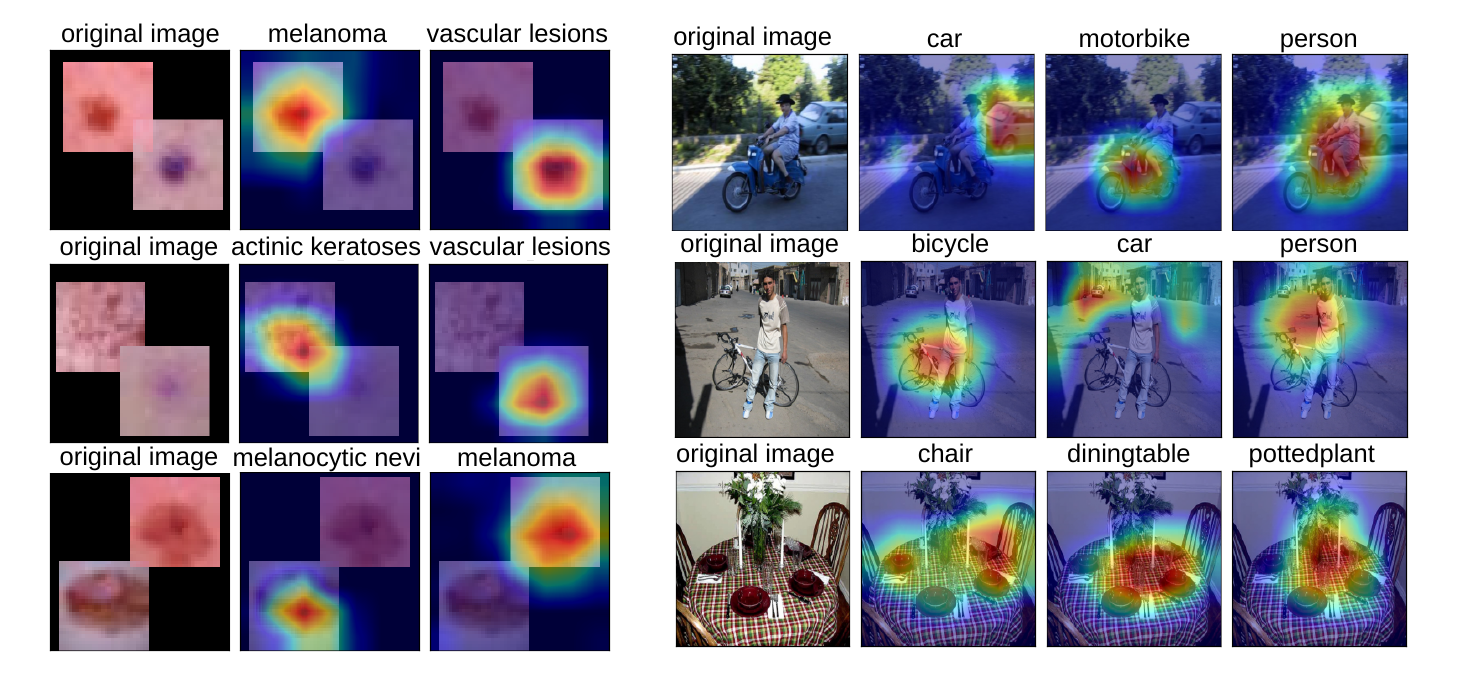} % left bottom right top
    \caption{
    Examples of Grad-CAM visualizations on the multi-label DermaMNIST and VOC datasets.
    Each subfigure shows an input image alongside class-specific Grad-CAM maps for all corresponding ground-truth labels.
    From the visualizations, FedNCA-ML captures class-specific evidence for each target class from the shared global image-level features. 
    Redder regions indicate stronger model responses for the corresponding class.
    }
    \label{fig: Ablation_attn_exmaple}
\end{figure}

\noindent\textbf{LADM Analysis.}
We also investigate the attention mechanism in the LADM module. 
As shown in Table~\ref{tab: Ablation_LADM_Derma-MNIST}, fixed, well-designed class-wise queries consistently outperform learnable queries across most metrics. 
This result is intuitive under label-skewed FL, where clients exhibit distinct local distributions. 
When both the queries and the classifier are learnable, each client can overfit to its local data and label relationships, increasing cross-client inconsistency and degrading the aggregated global model. 
In contrast, predefined and well-separated query embeddings provide a stable shared reference, encouraging more consistent optimization and improving robustness across non-IID clients.

\noindent\textbf{Visualization.}
To further analyse the model's behaviour, we present t-SNE visualisations of the test data in the latent feature space (Figure~\ref{fig: Ablation_Derma-MNIST_tsne}) and the pairwise cosine similarity between class-wise average features (Figure~\ref{fig: Ablation_Derma-MNIST_prototype}), obtained under different model architectures and training strategies on the multi-label DermaMNIST dataset.
As shown in Figure~\ref{fig: Ablation_Derma-MNIST_tsne_a}, when a conventional learnable fully connected (FC) classifier is used, the resulting feature representations exhibit poor clustering.
Notably, the model appears to rely on undesired information, grouping features by the number of labels per sample rather than solely by semantic content.
By incorporating LADM (Figure~\ref{fig: Ablation_Derma-MNIST_tsne_b}, \ref{fig: Ablation_Derma-MNIST_prototype_b}), which extracts single-class features, and further adding a predefined ETF classifier to regulate feature distribution across clients, the model learns to cluster features based on meaningful semantic attributes.
This results in improved clustering quality, as evidenced by the substantially reduced pairwise cosine similarity between class-wise average features, indicating enhanced inter-class separability and stronger discriminative capability.
Finally, incorporating additional regularization terms during training yields even more compact and semantically coherent feature clusters, with further reductions in prototype similarity (Figure~\ref{fig: Ablation_Derma-MNIST_tsne_c}, \ref{fig: Ablation_Derma-MNIST_prototype_c}).
In addition, Figure~\ref{fig: Ablation_attn_exmaple} presents Grad-CAM visualizations, demonstrating that FedNCA-ML consistently focuses on semantically relevant regions for each target class. 
This indicates that the model can localize class-specific evidence in the image, supporting accurate class-wise prediction.

\section{Conclusion}
\label{sec: conclusion}
This paper tackles the challenging problem of multi-label label-skewed FL.
This task is complicated by three intertwined factors: severe label imbalance, multi-label co-occurrence bias, and cross-client inconsistency in both label distributions and label relationships.
To address these issues, we propose FedNCA-ML, a pre-learning FL framework that structures and optimizes the latent feature space with a Neural Collapse–inspired geometry, promoting cross-client representation consistency under non-IID data. 
FedNCA-ML integrates a class-wise feature extraction module with a predefined ETF as a shared geometric reference, inducing NC-style clustering in multi-label settings and guiding clients toward a coherent optimization objective. 
We further introduce two regularization losses to suppress noisy signals and encourage compact, well-separated class-wise clustering in the latent space.
Experiments on five datasets under nine different FL settings demonstrate the effectiveness and robustness of the proposed method.

\section*{Acknowledgments}
This work was supported by the UKRI grant EP/X040186/1 (Turing AI Fellowship). 
This work was also partly supported by the InnoHK-funded Hong Kong Centre for Cerebrocardiovascular Health Engineering (COCHE) Project 2.1 (Cardiovascular risks in early life and fetal echocardiography).

% ---------------------------------------------------------------
% Bibliography
\clearpage
\bibliographystyle{unsrt}
\bibliography{main}

% ---------------------------------------------------------------
% Appendix / Supplementary
\clearpage
\appendix
\clearpage
\appendix
\section*{Appendix}
\addcontentsline{toc}{section}{Appendix}

\section{Dataset}
\label{sup_sec: dataset}
In this section, we introduce the datasets used in our experiments. 
To evaluate both the effectiveness and real-world applicability of the proposed method, we conduct experiments on datasets from both general computer vision (CV) as well as medical imaging domains. 
Specifically, we use CIFAR-10~\cite{krizhevsky2009learning}, PASCAL VOC~\cite{everingham2010pascal}, and MS COCO~\cite{lin2014microsoft} as general CV benchmarks, and DermaMNIST~\cite{yang2023medmnist} and ChestX-ray14~\cite{wang2017chestxray} as medical imaging datasets.
To simulate non-IID federated learning (FL) settings,  we partition the data using a Dirichlet distribution, with the concentration parameter $\beta$ controlling the degree of heterogeneity. 
To further model missing-class scenarios, we constrain the number of classes available to each client using the class presence ratio $\gamma$, which specifies the proportion of total classes present locally. 
Dataset-specific settings are detailed below.

\begin{figure}[tbp]
    \centering
    \begin{subfigure}[t]{0.49\linewidth}
        \centering
        \includegraphics[width=\linewidth, trim={0.7cm 0.2cm 0.7cm 0.2cm}, clip]{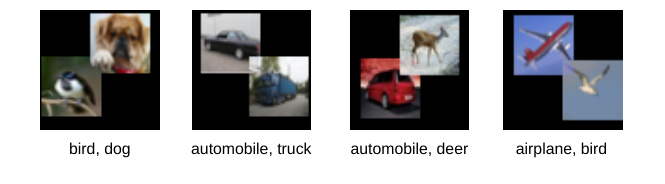}
        \caption{Examples for the multi-label CIFAR-10 dataset.}
        \label{fig:multi_CIFAR10_example}
    \end{subfigure}
    \hfill   
    \begin{subfigure}[t]{0.49\linewidth}
        \centering
        \includegraphics[width=\linewidth, trim={0.7cm 0.2cm 0.7cm 0.2cm}, clip]{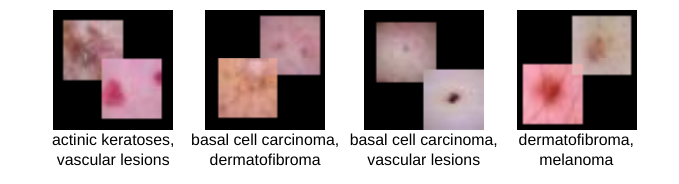}
        \caption{Examples for the multi-label DermaMNIST dataset.}
        \label{fig:multi_DermaMNIST_example}
    \end{subfigure}
    \caption{Examples for the multi-label CIFAR-10 and DermaMNIST datasets. 
    Each subfigure displays a composite image along with its corresponding set of labels.}
    \label{fig:multi_dataset_examples}
\end{figure}

\begin{figure*}[p]
    \centering
    \begin{subfigure}[c]{0.14\linewidth}
        \centering
        \includegraphics[width=\linewidth, trim={0cm 0cm 3.5cm 0cm},clip]{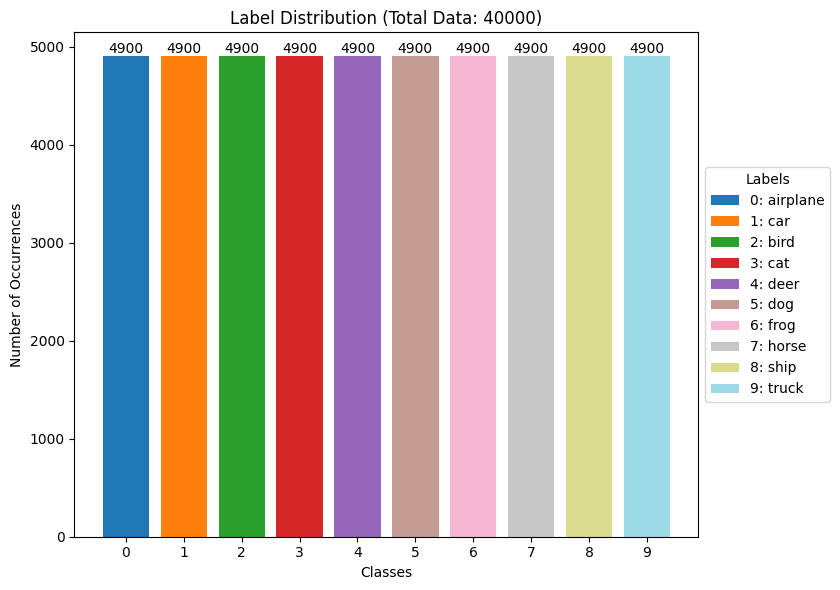} % left bottom right top 
        \caption{Overall training data distribution.}
    \end{subfigure}
    \hfill
    \begin{subfigure}[c]{0.41\linewidth}
        \centering
        \includegraphics[width=\linewidth, trim={0cm 0cm 0cm 0.86cm},clip]{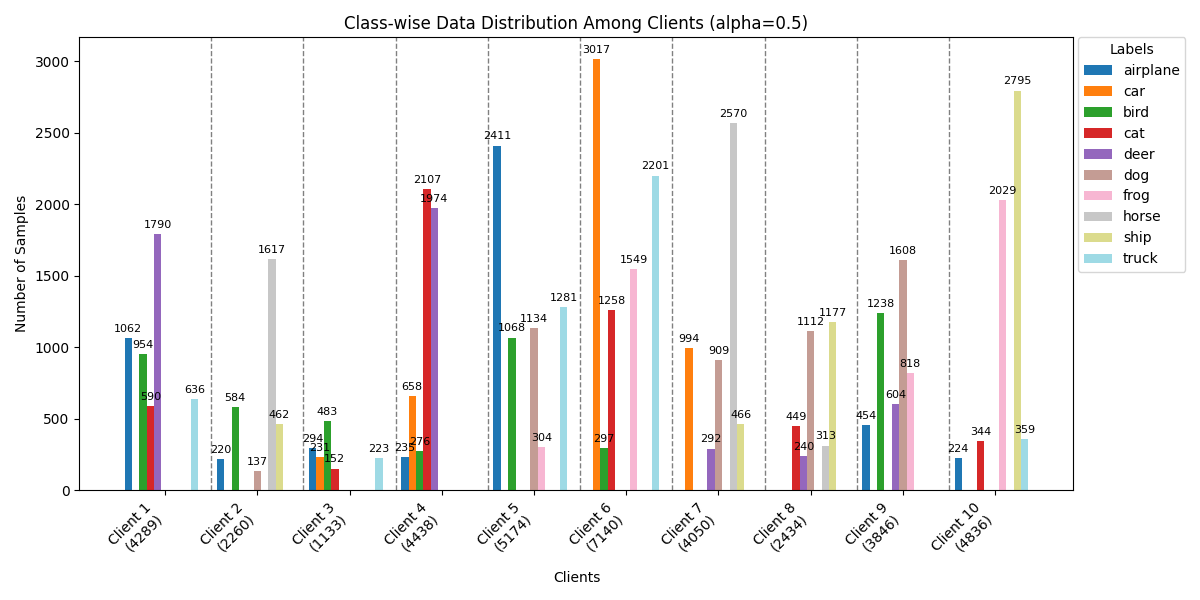}
        \caption{Per-client training data distribution with $\beta = 0.5$ and $\gamma = 0.5$.}
    \end{subfigure}
    \hfill
    \begin{subfigure}[c]{0.41\linewidth}
        \centering
        \includegraphics[width=\linewidth, trim={0cm 0cm 0cm 0.86cm},clip]{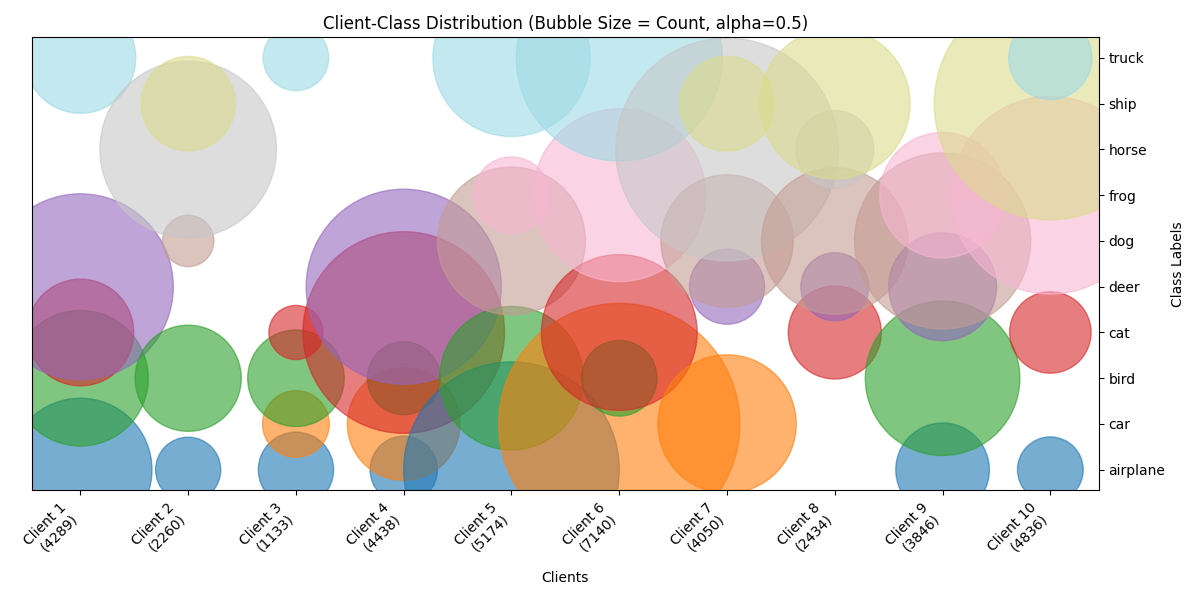}
        \caption{Per-client training data bubble chart under $\beta = 0.5$ and $\gamma = 0.5$.}
    \end{subfigure}
    \centering
    \begin{subfigure}[c]{0.14\linewidth}
        \centering
        \includegraphics[width=\linewidth, trim={0cm 0cm 3.5cm 0cm},clip]{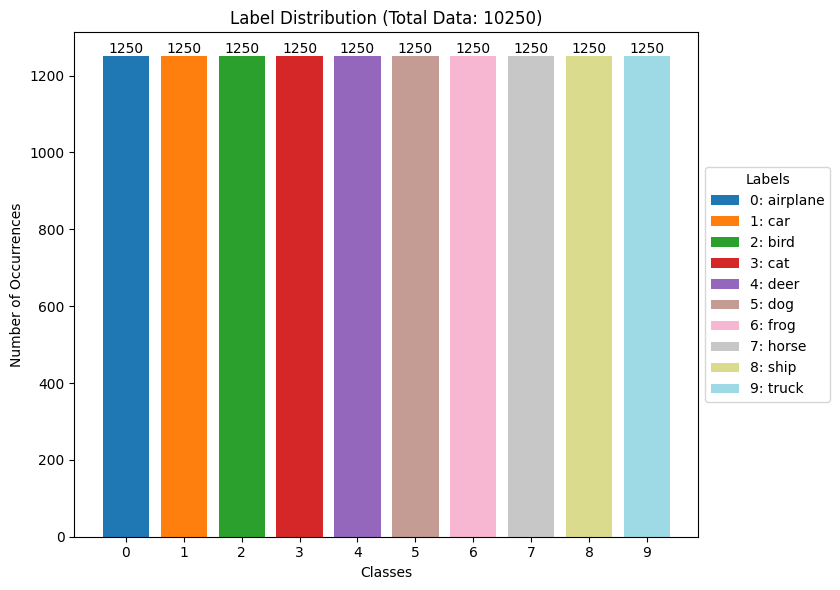} % left bottom right top
        \caption{Testing data distribution.}
    \end{subfigure}
    \hfill
    \begin{subfigure}[c]{0.41\linewidth}
        \centering
        \includegraphics[width=\linewidth, trim={0cm 0cm 0cm 0.86cm},clip]{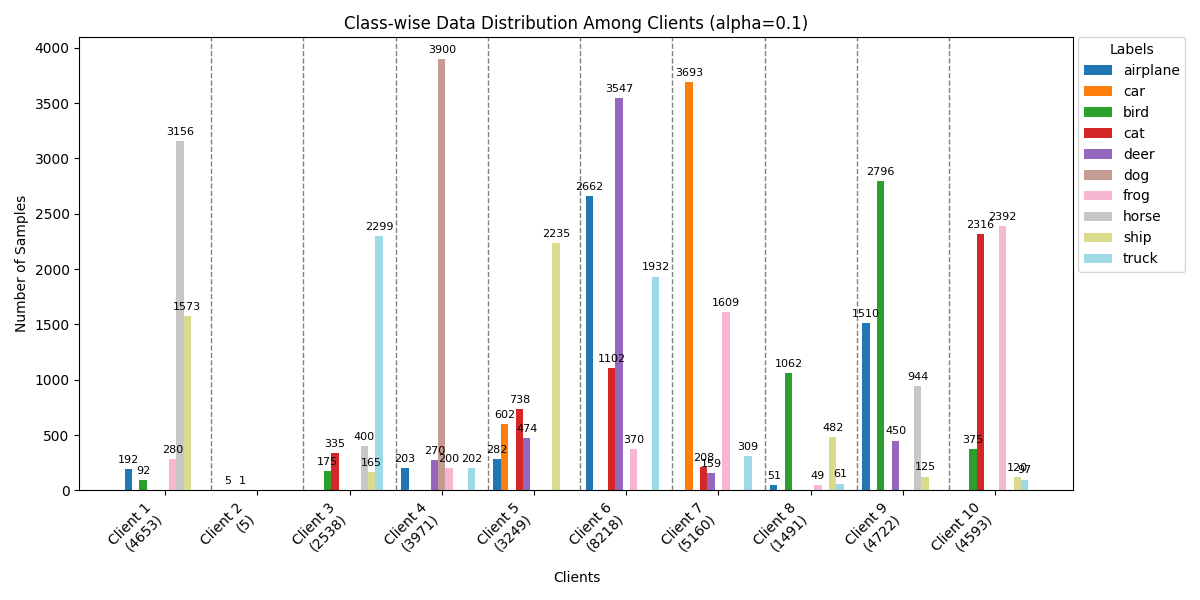}
        \caption{Per-client training data distribution under $\beta = 0.1$ and $\gamma = 0.5$.}
    \end{subfigure}
    \hfill
    \begin{subfigure}[c]{0.41\linewidth}
        \centering
        \includegraphics[width=\linewidth, trim={0cm 0cm 0cm 0.86cm},clip]{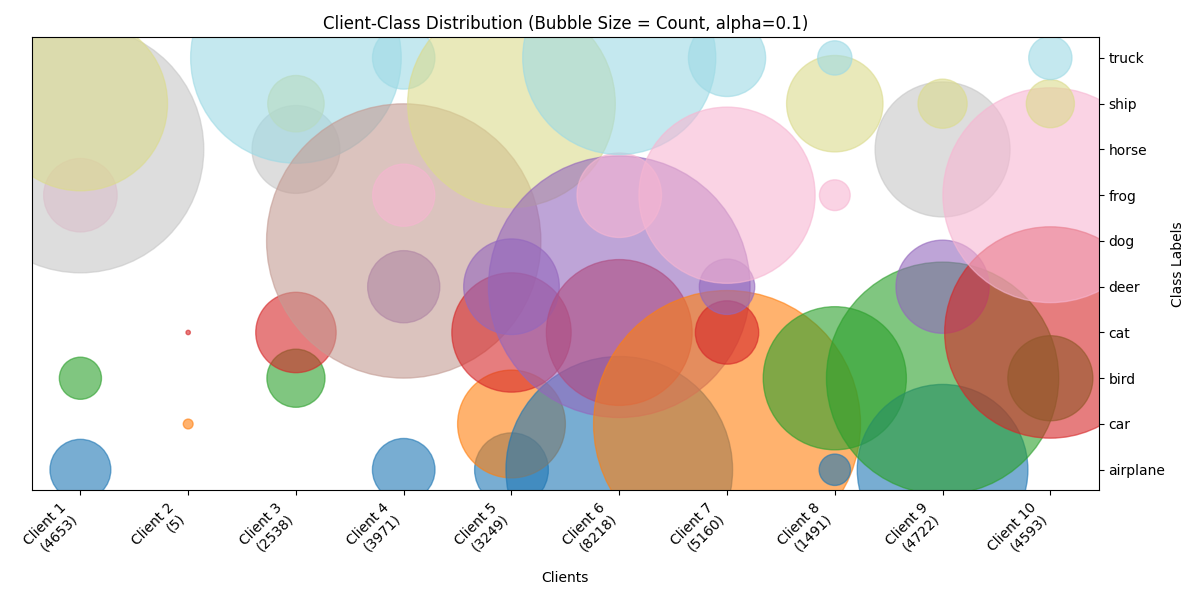}
        \caption{Per-client training data bubble chart under $\beta = 0.1$ and $\gamma = 0.5$.}
    \end{subfigure}
    \caption{
    Distribution of data across local clients in the CIFAR-10~\cite{krizhevsky2009learning} experiments. 
    The class presence ratio ($\gamma$) is set to 0.5 ($\leq$ 5 of 10 classes per client).
    Non-IID client distributions are simulated using the Dirichlet factor ($\beta$).
    }
    \label{fig:CIFAR10_data_distribution}
% ================================================================================================
\vspace{2em}
    \centering
    \begin{subfigure}[c]{0.14\linewidth}
        \centering
        \includegraphics[width=\linewidth, trim={0cm 0cm 5cm 0cm},clip]{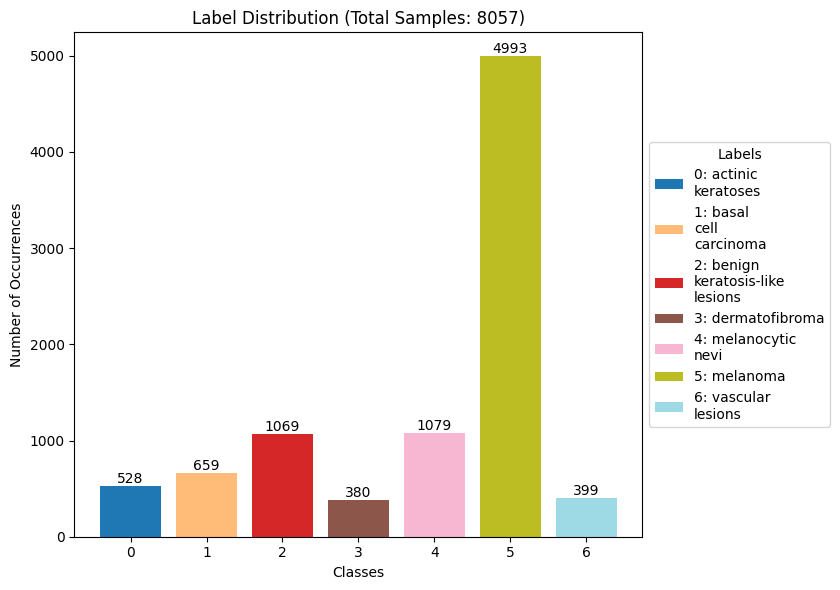} % left bottom right top
        \caption{Overall training data distribution.}
    \end{subfigure}
    \begin{subfigure}[c]{0.41\linewidth}
        \centering
        \includegraphics[width=\linewidth, trim={0cm 0cm 0cm 0.86cm},clip]{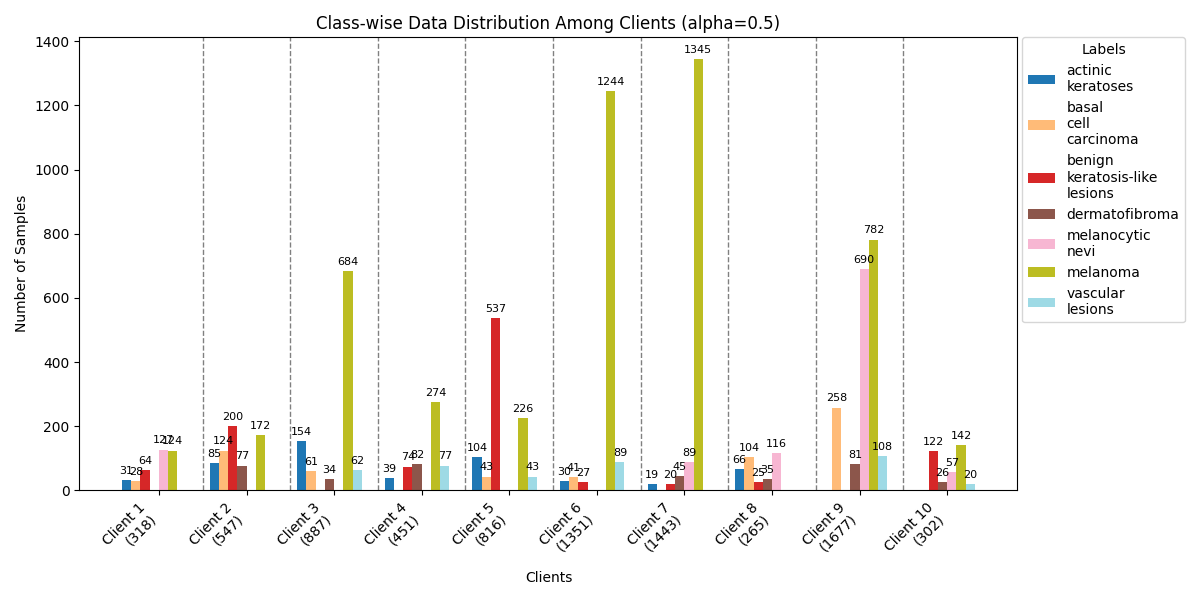} % left bottom right top
        \caption{Per-client training data distribution under $\beta = 0.5$ and $\gamma = 0.71$.}
    \end{subfigure}
    \begin{subfigure}[c]{0.41\linewidth}
        \centering
        \includegraphics[width=\linewidth, trim={0cm 0cm 0cm 0.86cm},clip]{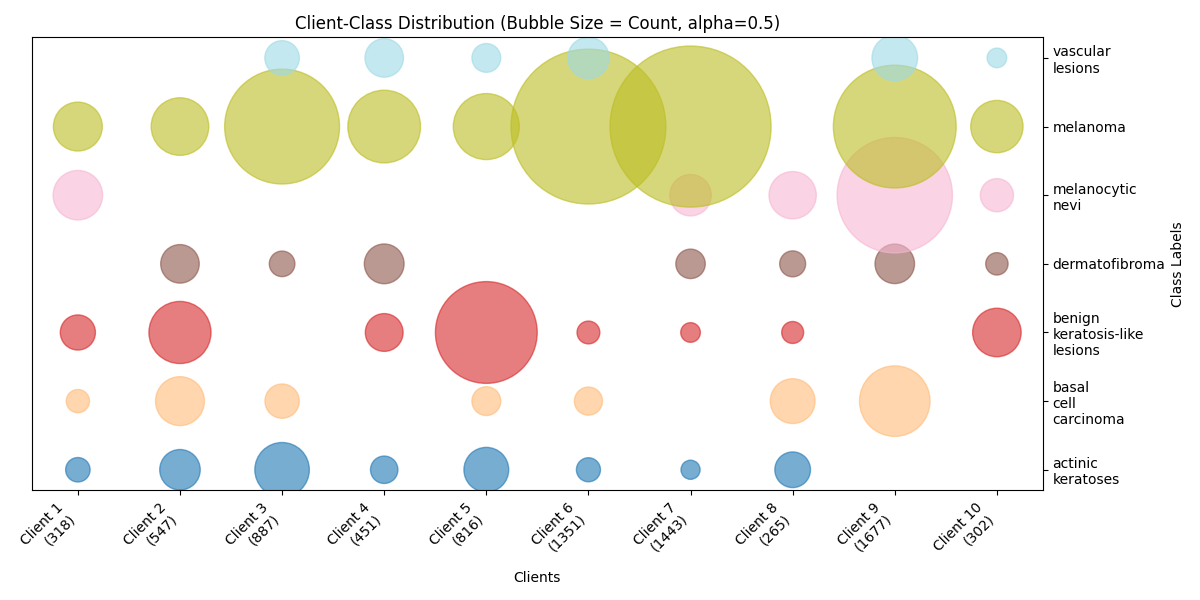} % left bottom right top
        \caption{Per-client training data bubble chart under $\beta = 0.5$ and $\gamma = 0.71$.}
    \end{subfigure}
    \begin{subfigure}[c]{0.14\linewidth}
        \centering
        \includegraphics[width=\linewidth, trim={0cm 0cm 5cm 0cm},clip]{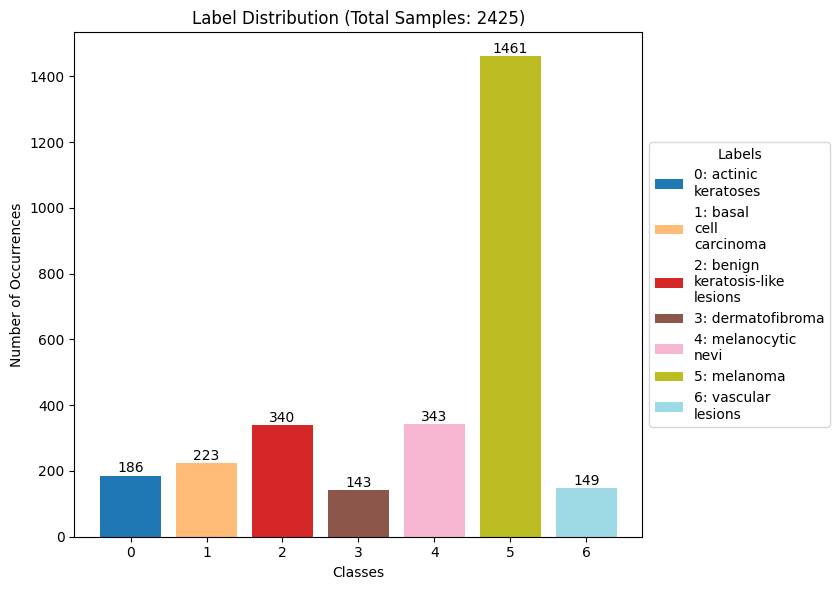} % left bottom right top
        \caption{Testing data distribution.}
    \end{subfigure}
    \begin{subfigure}[c]{0.41\linewidth}
        \centering
        \includegraphics[width=\linewidth, trim={0cm 0cm 0cm 0.86cm},clip]{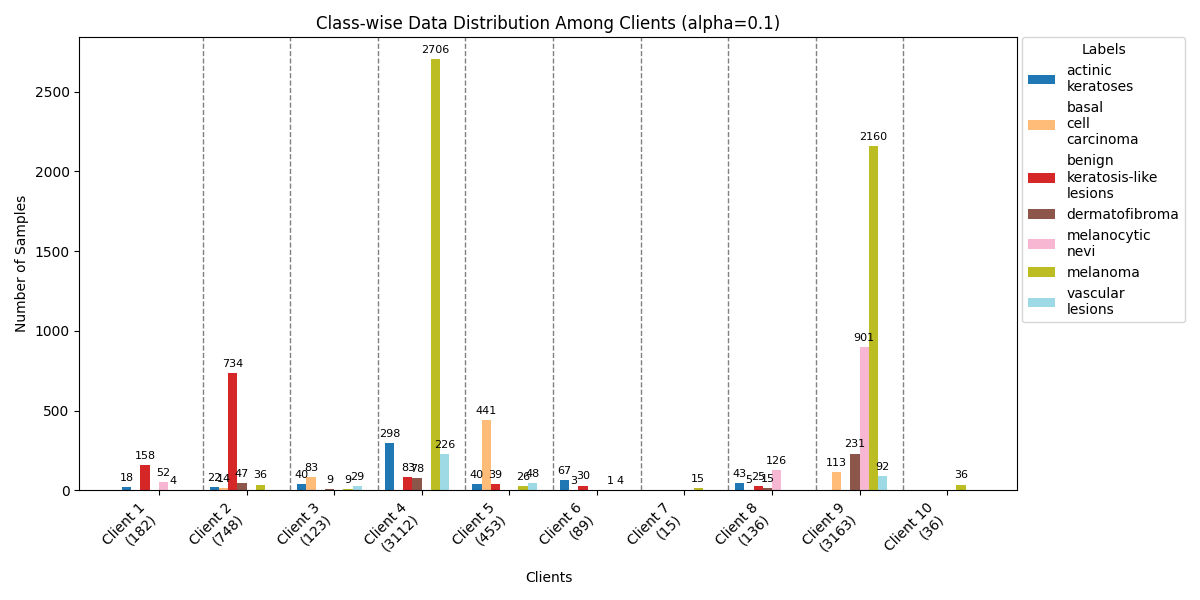} % left bottom right top
        \caption{Per-client training data distribution under $\beta = 0.1$ and $\gamma = 0.71$.}
    \end{subfigure}
    \begin{subfigure}[c]{0.41\linewidth}
        \centering
        \includegraphics[width=\linewidth, trim={0cm 0cm 0cm 0.86cm},clip]{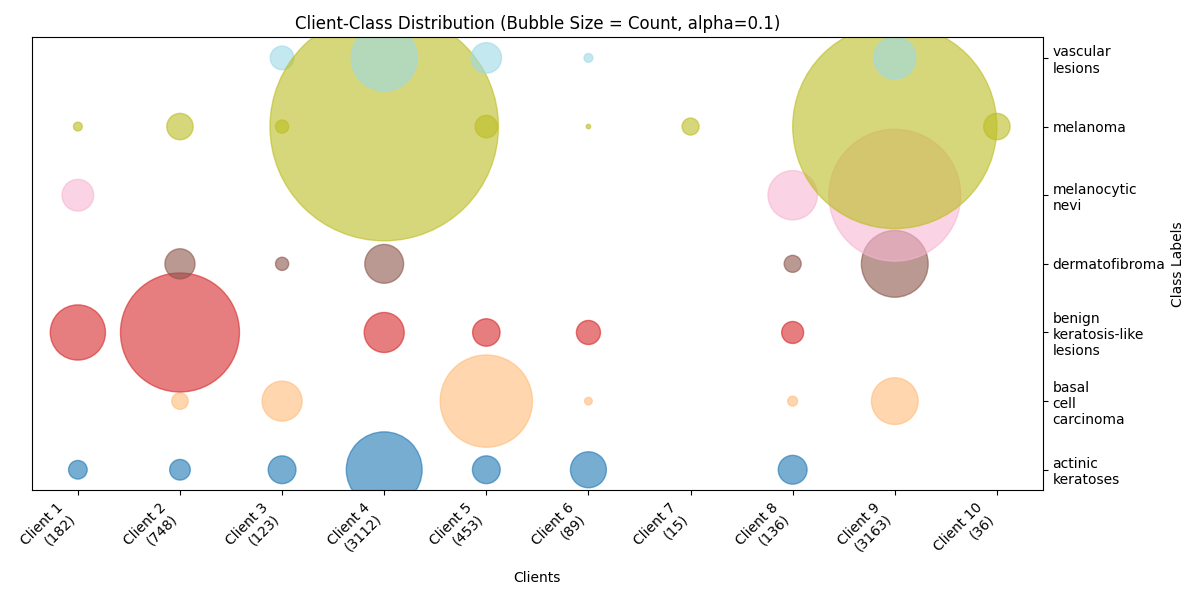} % left bottom right top
        \caption{Per-client training data bubble chart under $\beta = 0.1$ and $\gamma = 0.71$.}
    \end{subfigure}
    \caption{
    Distribution of data across local clients in the DermaMNIST~\cite{yang2023medmnist} experiments.
    The class presence ratio ($\gamma$) is set to 0.71 ($\leq$ 5 of 7 classes per client).
    Non-IID client distributions are simulated using the Dirichlet factor ($\beta$).
    }
	\label{fig:DermaMNIST_data_distribution}
\end{figure*}

\noindent \textbf{CIFAR-10}~\cite{krizhevsky2009learning} is a widely used benchmark dataset in CV.
It comprises 10 classes: \textit{airplane}, \textit{automobile}, \textit{bird}, \textit{cat}, \textit{deer}, \textit{dog}, \textit{frog}, \textit{horse}, \textit{ship}, and \textit{truck}.
To adapt CIFAR-10 to the multi-label setting, following \cite{li2023neural}, we construct composite samples by combining multiple original images into a single image, with the associated class labels forming the multi-label ground truth. 
Specifically, we retain a portion of the original single-label images and augment the dataset by generating an equal number of synthetic samples for every possible pairwise class combination. 
Each composite sample is created by randomly selecting two images from different classes and merging them into a single image. 
Examples of the resulting multi-label composite samples are shown in Figure~\ref{fig:multi_CIFAR10_example}. 
We evaluate the proposed method on the resulting multi-label CIFAR-10 dataset under two FL settings: $\beta = 0.5,\ \gamma = 0.5$ and $\beta = 0.1,\ \gamma = 0.5$. 
The corresponding class-wise data distributions across clients are illustrated in Figure~\ref{fig:CIFAR10_data_distribution}.

\noindent \textbf{DermaMNIST}~\cite{yang2023medmnist} is a skin lesion classification dataset.
It consists of 7 diagnostic categories: \textit{actinic keratoses}, \textit{basal cell carcinoma}, \textit{benign keratosis-like lesions}, \textit{dermatofibroma}, \textit{melanocytic nevi}, \textit{melanoma}, and \textit{vascular lesions}.
To adapt DermaMNIST to the multi-label setting, we adopt a strategy similar to that used for multi-label CIFAR-10. Given the long-tailed nature of the original dataset, we preserve its intrinsic class distribution by retaining all original single-label samples. 
We then augment the dataset by generating an equal number of synthetic samples for each possible pairwise label combination. 
Examples of the resulting composite multi-label samples are shown in Figure~\ref{fig:multi_DermaMNIST_example}. 
The final dataset consists of the complete set of original samples together with the newly generated multi-label samples. 
We evaluate the proposed method on this multi-label DermaMNIST dataset under two FL settings: $\beta = 0.5, \gamma = 0.71$ and $\beta = 0.1, \gamma = 0.71$. 
The corresponding class-wise data distributions across clients are illustrated in Figure~\ref{fig:DermaMNIST_data_distribution}.

\begin{figure*}[p]
    \centering
    \begin{subfigure}[c]{0.14\linewidth}
        \centering
        \includegraphics[width=\linewidth, trim={0cm 0cm 4.2cm 0cm},clip]{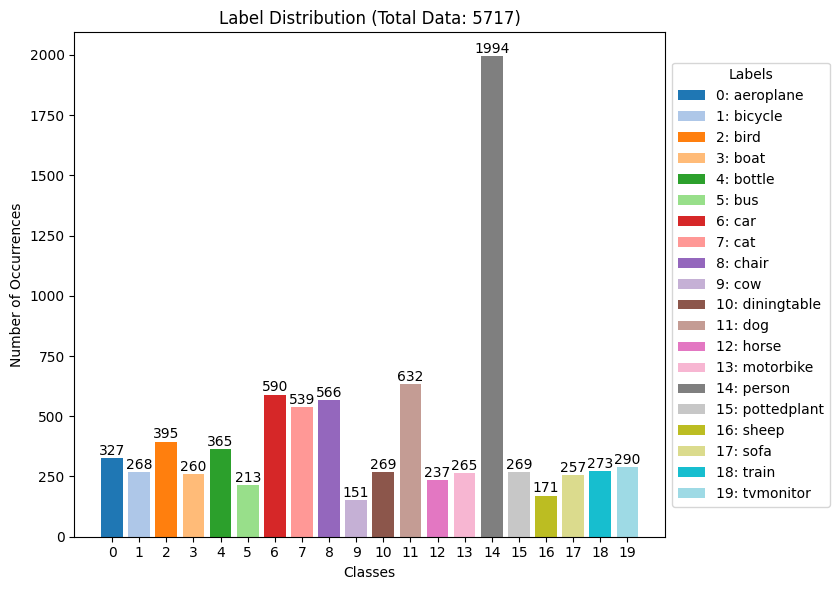} % left bottom right top
        \caption{Overall training data distribution.}
    \end{subfigure}
    \begin{subfigure}[c]{0.41\linewidth}
        \centering
        \includegraphics[width=\linewidth, trim={0cm 0cm 0cm 0.86cm},clip]{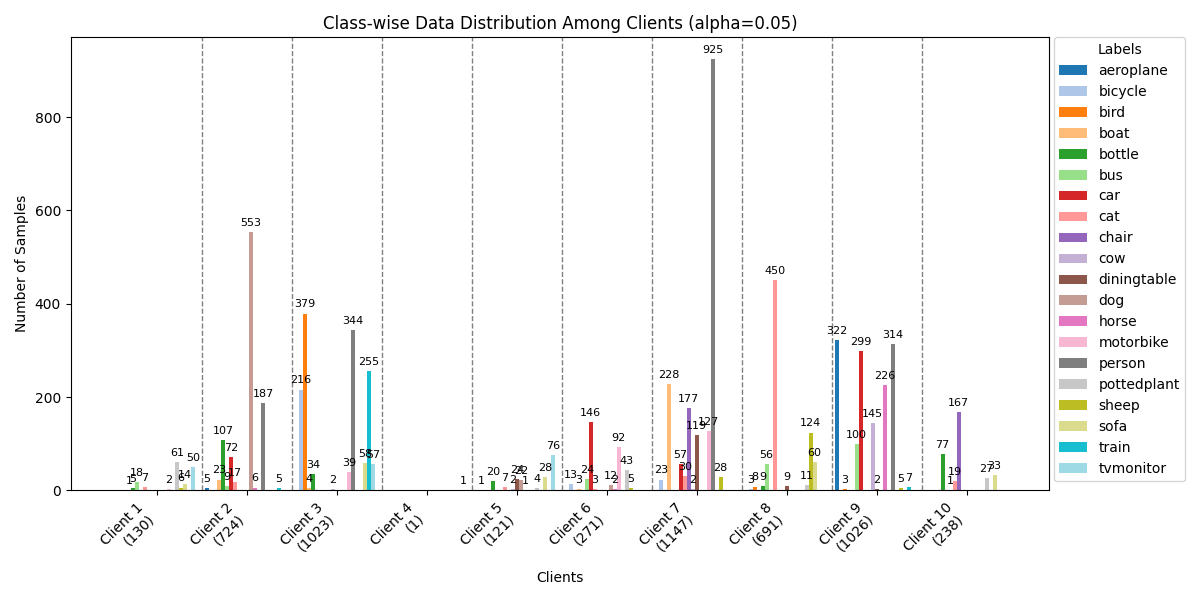} % left bottom right top
        \caption{Per-client training data distribution under $\beta = 0.05$ and $\gamma = 0.5$.}
    \end{subfigure}
    \begin{subfigure}[c]{0.41\linewidth}
        \centering
        \includegraphics[width=\linewidth, trim={0cm 0cm 0cm 0.86cm},clip]{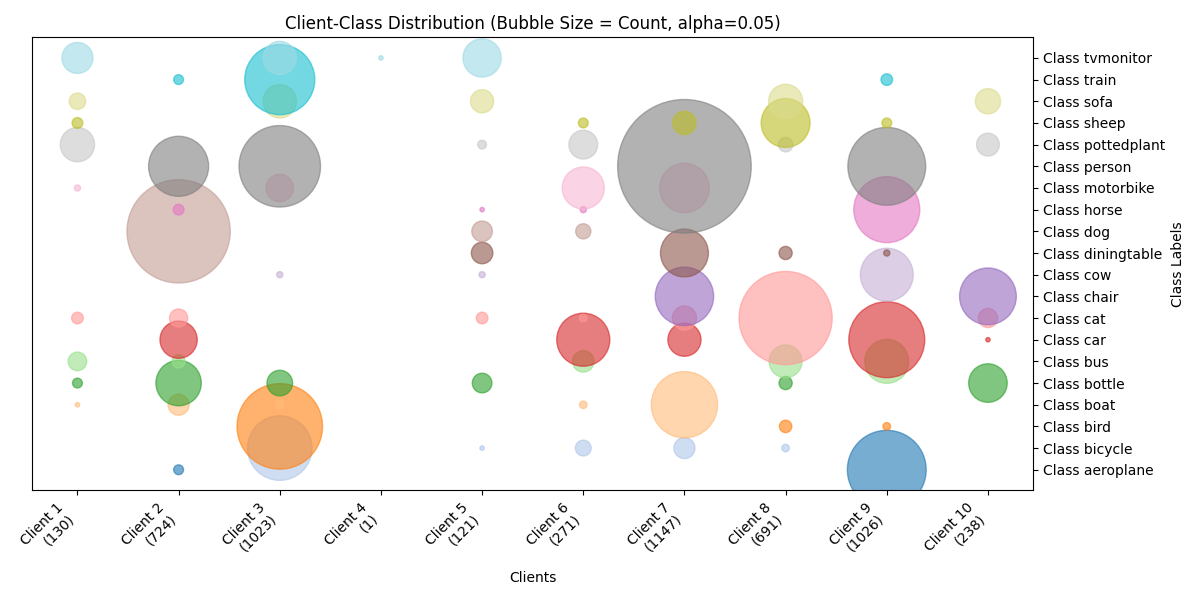} % left bottom right top
        \caption{Per-client training data bubble chart under $\beta = 0.05$ and $\gamma = 0.5$.}
    \end{subfigure}
    \begin{subfigure}[c]{0.14\linewidth}
        \centering
        \includegraphics[width=\linewidth, trim={0cm 0cm 4.2cm 0cm},clip]{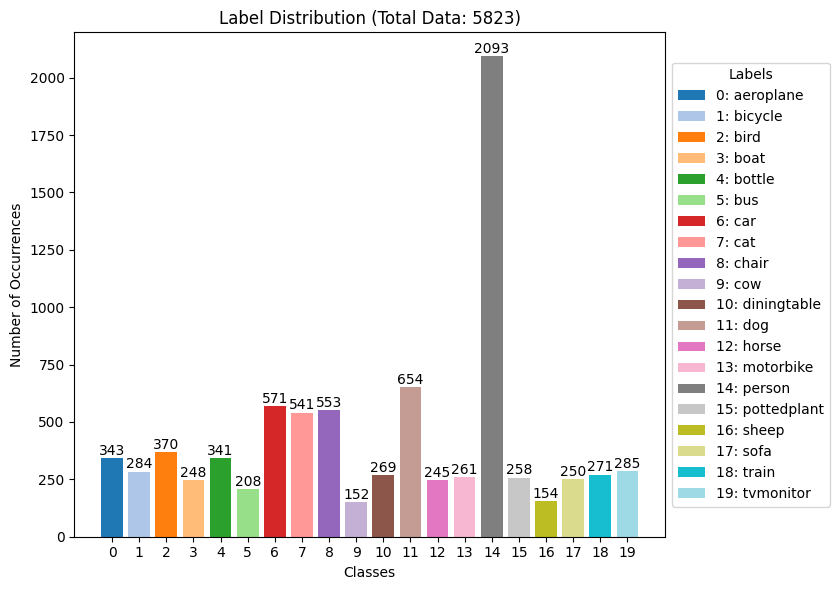} % left bottom right top
        \caption{Testing data distribution.}
    \end{subfigure}
    \begin{subfigure}[c]{0.41\linewidth}
        \centering
        \includegraphics[width=\linewidth, trim={0cm 0cm 0.86cm 0cm},clip]{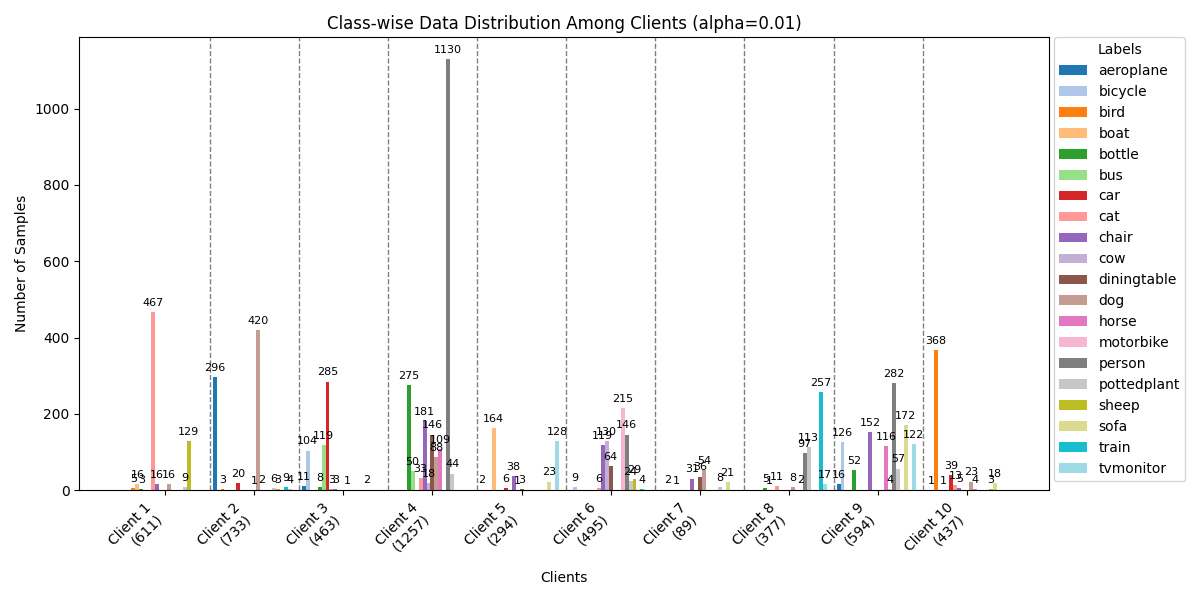} % left bottom right top
        \caption{Per-client training data distribution under $\beta = 0.01$ and $\gamma = 0.5$.}
    \end{subfigure}
    \begin{subfigure}[c]{0.41\linewidth}
        \centering
        \includegraphics[width=\linewidth, trim={0cm 0cm 0cm 0.86cm},clip]{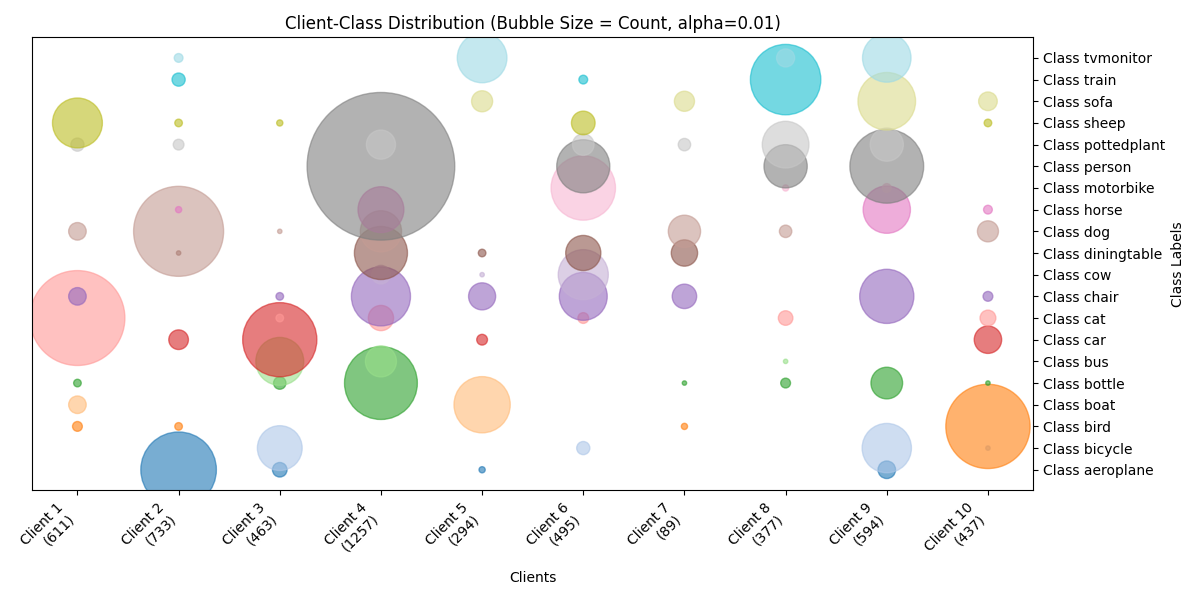} % left bottom right top
        \caption{Per-client training data bubble chart under $\beta = 0.01$ and $\gamma = 0.5$.}
    \end{subfigure}
    \caption{
    Distribution of data across local clients in the PASCAL VOC~\cite{everingham2010pascal} experiments.
    The class presence ratio ($\gamma$) is set to 0.5 ($\leq$ 10 of 20 classes per client).
    Non-IID client distributions are simulated using the Dirichlet factor ($\beta$).
    }
	\label{fig:VOC_data_distribution}

% ================================================================================================
\vspace{2em}
    \centering
    \begin{subfigure}[c]{0.14\linewidth}
        \centering
        \includegraphics[width=\linewidth, trim={0cm 0cm 5.5cm 0cm},clip]{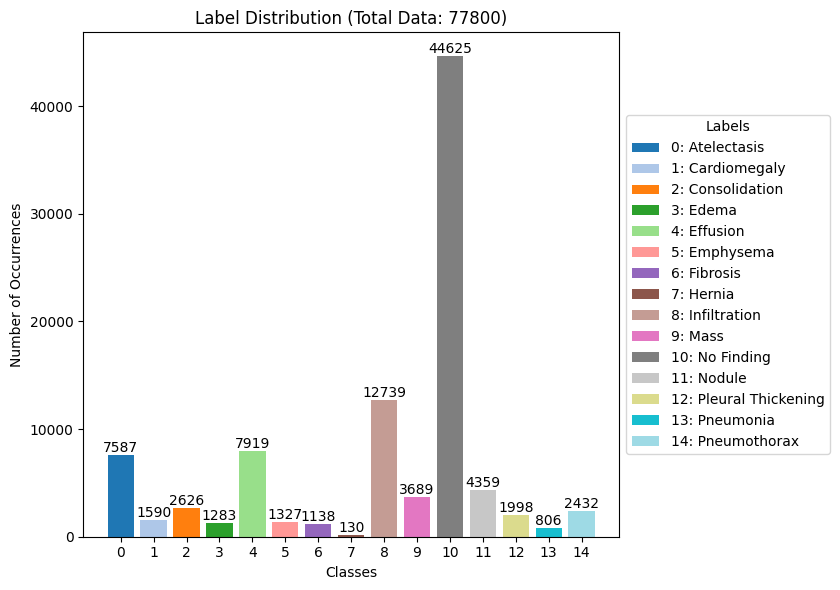} % left bottom right top
        \caption{Overall training data distribution.}
    \end{subfigure}
    \begin{subfigure}[c]{0.37\linewidth}
        \centering
        \includegraphics[width=\linewidth, trim={0cm 0cm 0cm 0.86cm},clip]{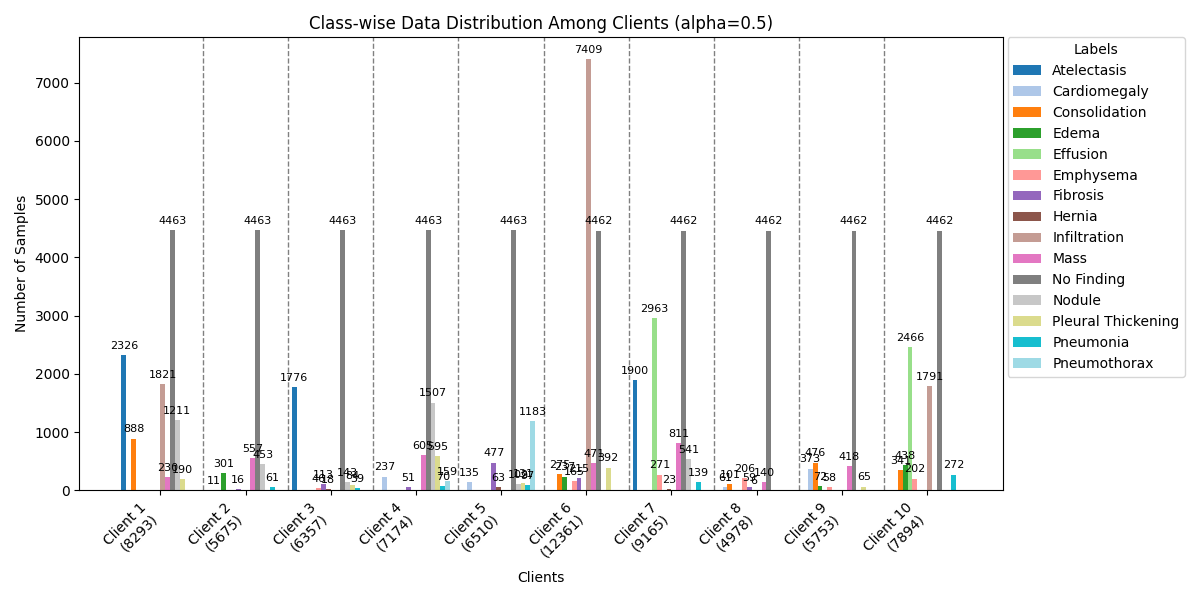} % left bottom right top
        \caption{Per-client training data distribution under $\beta = 0.5$ and $\gamma = 0.5$.}
    \end{subfigure}
    \begin{subfigure}[c]{0.41\linewidth}
        \centering
        \includegraphics[width=\linewidth, trim={0cm 0cm 0cm 0.86cm},clip]{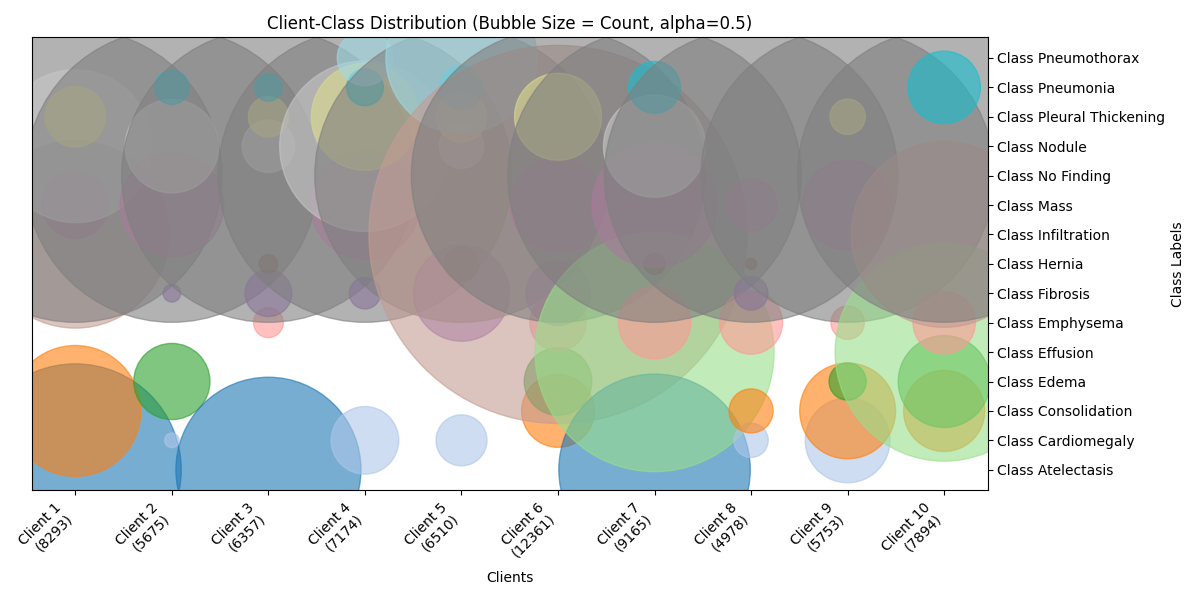} % left bottom right top
        \caption{Per-client training data bubble chart under $\beta = 0.5$ and $\gamma = 0.5$.}
    \end{subfigure}
    \begin{subfigure}[c]{0.14\linewidth}
        \centering
        \includegraphics[width=\linewidth, trim={0cm 0cm 5.5cm 0cm},clip]{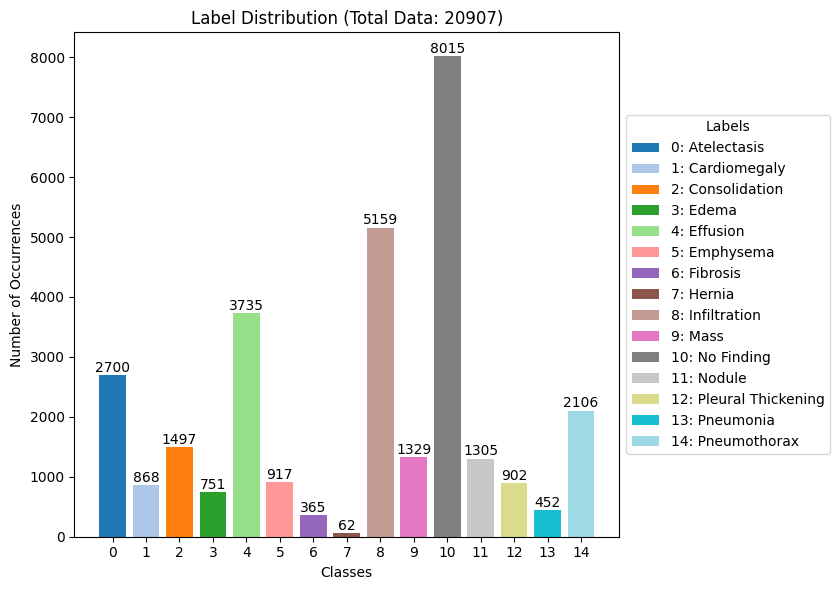} % left bottom right top
        \caption{Testing data distribution.}
    \end{subfigure}
    \begin{subfigure}[c]{0.41\linewidth}
        \centering
        \includegraphics[width=\linewidth, trim={0cm 0cm 0cm 0.86cm},clip]{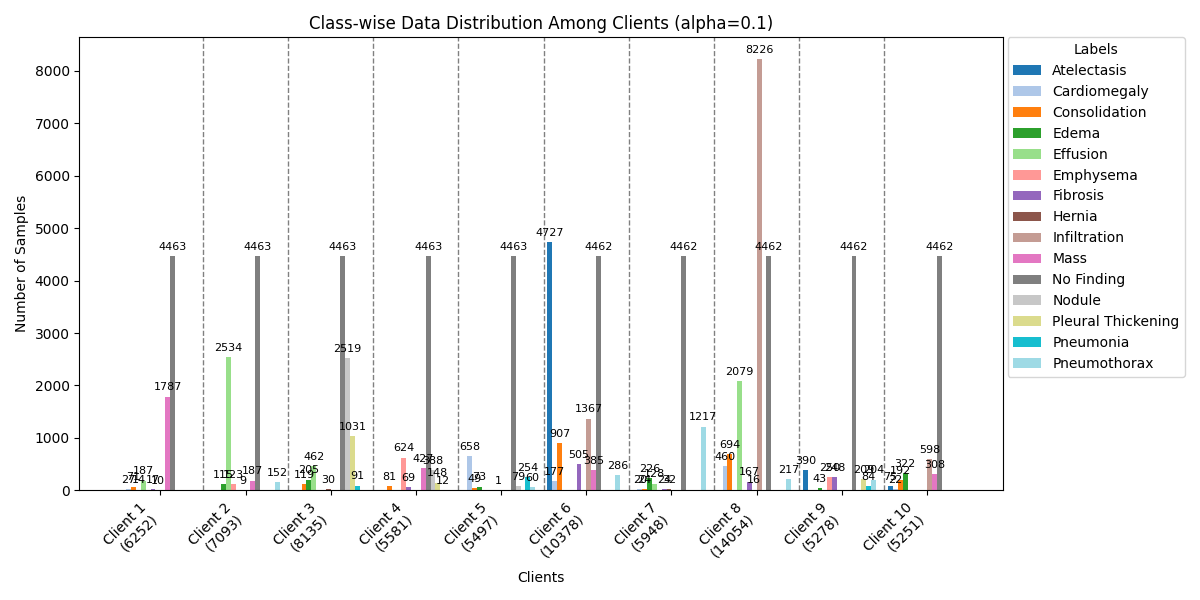} % left bottom right top
        \caption{Per-client training data distribution under $\beta = 0.1$ and $\gamma = 0.5$.}
    \end{subfigure}
    \begin{subfigure}[c]{0.41\linewidth}
        \centering
        \includegraphics[width=\linewidth, trim={0cm 0cm 0cm 0.86cm},clip]{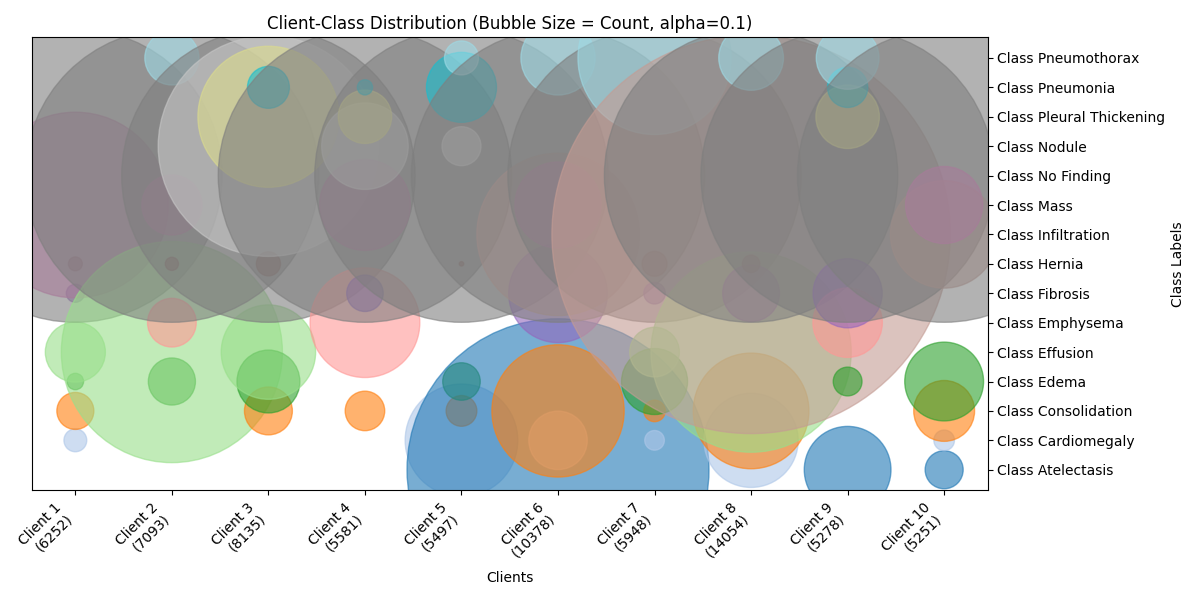} % left bottom right top
        \caption{Per-client training data bubble chart under $\beta = 0.1$ and $\gamma = 0.5$.}
    \end{subfigure}
    \caption{
    Distribution of data across local clients in the ChestX-ray14~\cite{wang2017chestxray} experiments.
    The ChestX-ray14 dataset contains 14 thoracic disease categories and an additional ``No Finding'' label. 
    Since a large portion of the dataset (57\% of the training data) is ``No Finding'' samples (i.e., negative cases with all-zero labels), we distribute these samples evenly across all clients to reflect a realistic clinical scenario in which healthy cases are prevalent.
    The class presence ratio ($\gamma$) is set to 0.5 ($\leq$ 7 of 14 disease classes per client).
    Non-IID client distributions are simulated using the Dirichlet factor ($\beta$).
    }
	\label{fig:Xray_data_distribution}
\end{figure*}

\noindent \textbf{PASCAL VOC}~\cite{everingham2010pascal} is a widely used benchmark dataset in CV. 
It contains approximately 11,500 images, each annotated with one or more object categories from a predefined set of 20 classes. 
These categories include: \textit{aeroplane}, \textit{bicycle}, \textit{bird}, \textit{boat}, \textit{bottle}, \textit{bus}, \textit{car}, \textit{cat}, \textit{chair}, \textit{cow}, \textit{dining table}, \textit{dog}, \textit{horse}, \textit{motorbike}, \textit{person}, \textit{potted plant}, \textit{sheep}, \textit{sofa}, \textit{train}, and \textit{TV monitor}.
This dataset presents a challenging multi-label classification task due to substantial class imbalance, high intra-class variability, and frequent inter-class co-occurrence.
For instance, the \textit{person} class commonly appears alongside many other object categories.
Under label-skewed FL settings, these challenges are further amplified by non-IID client distributions and inconsistent label co-occurrence patterns across clients. 
We evaluate the proposed method on PASCAL VOC under two FL settings: $\beta = 0.05, \gamma = 0.5$ and $\beta = 0.01, \gamma = 0.5$. 
The resulting class-wise data distributions across clients are shown in Figure~\ref{fig:VOC_data_distribution}.

\begin{figure}[t]
    \centering
    \begin{minipage}[c]{0.6\linewidth}
        \centering
        \begin{subfigure}[c]{\linewidth}
            \centering
            \includegraphics[width=\linewidth, trim={0.2cm 0cm 0.2cm 0cm}, clip]{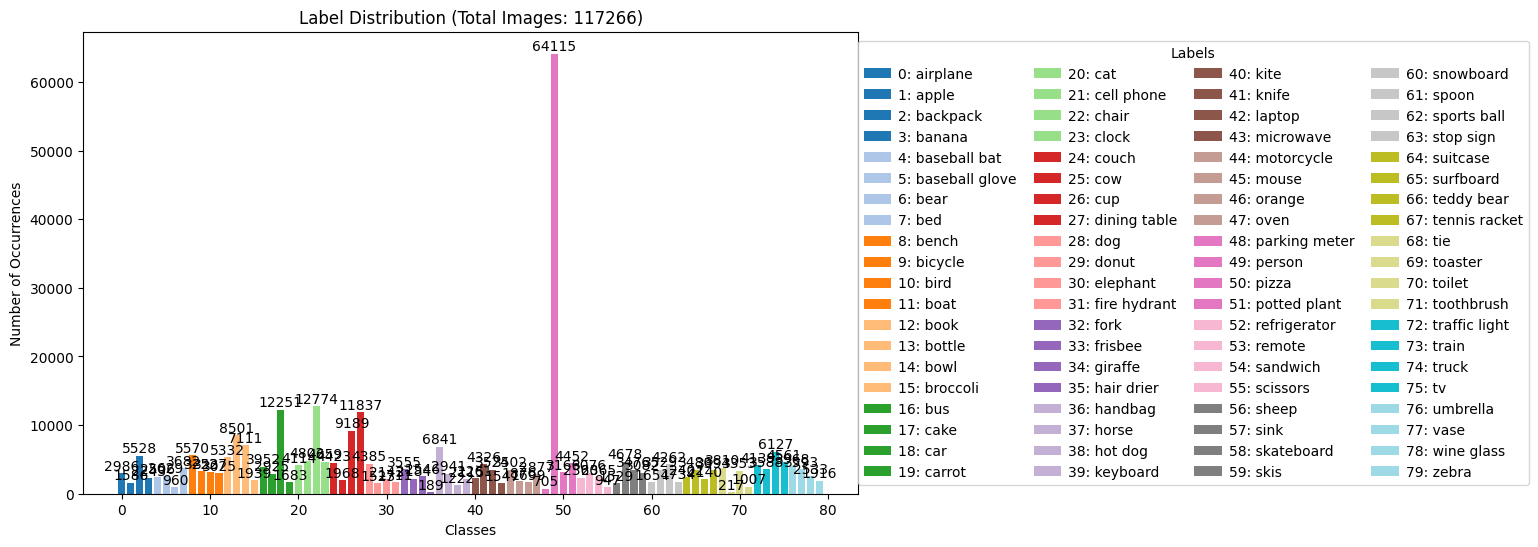}
            \caption{Overall training data distribution.}
            \vspace{2em}
        \end{subfigure}
        \begin{subfigure}[c]{\linewidth}
            \centering
            \includegraphics[width=\linewidth, trim={0.2cm 0cm 0.2cm 0cm}, clip]{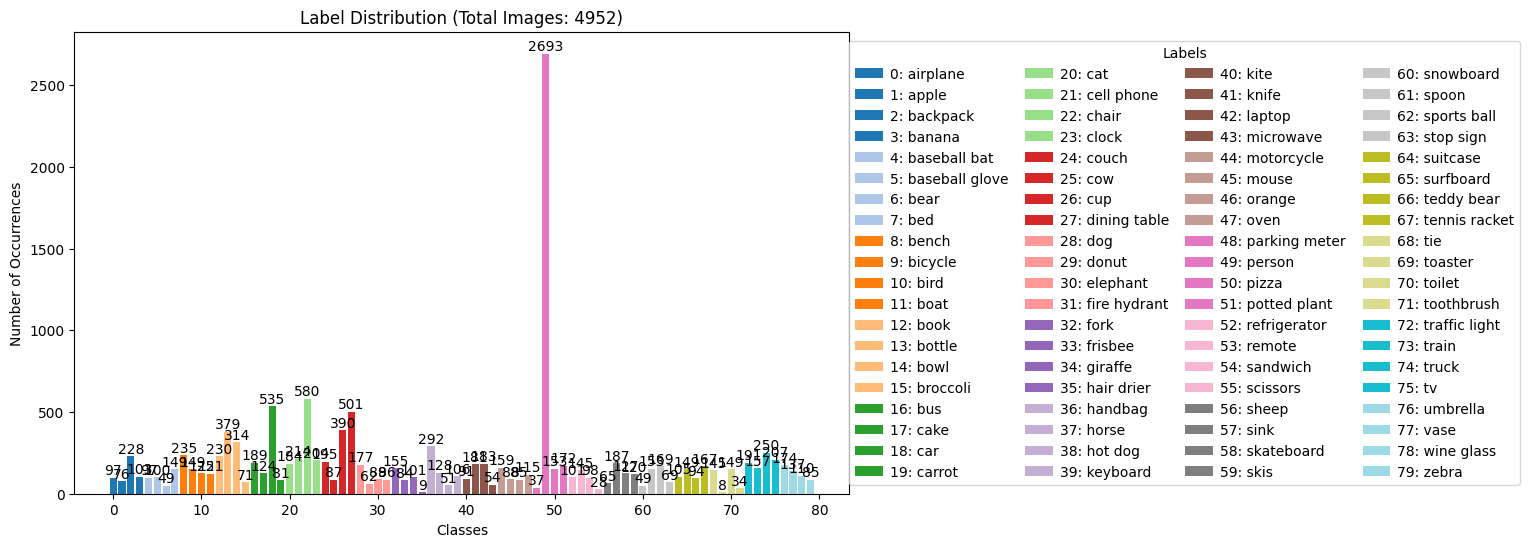}
            \caption{Testing data distribution.}
        \end{subfigure}
    \end{minipage}
    \hfill
    \begin{minipage}[c]{0.39\linewidth}
        \centering
        \begin{subfigure}[c]{\linewidth}
            \centering
            \includegraphics[width=0.9\linewidth,trim={0.3cm 0cm 0.3cm 0cm}, clip, keepaspectratio]{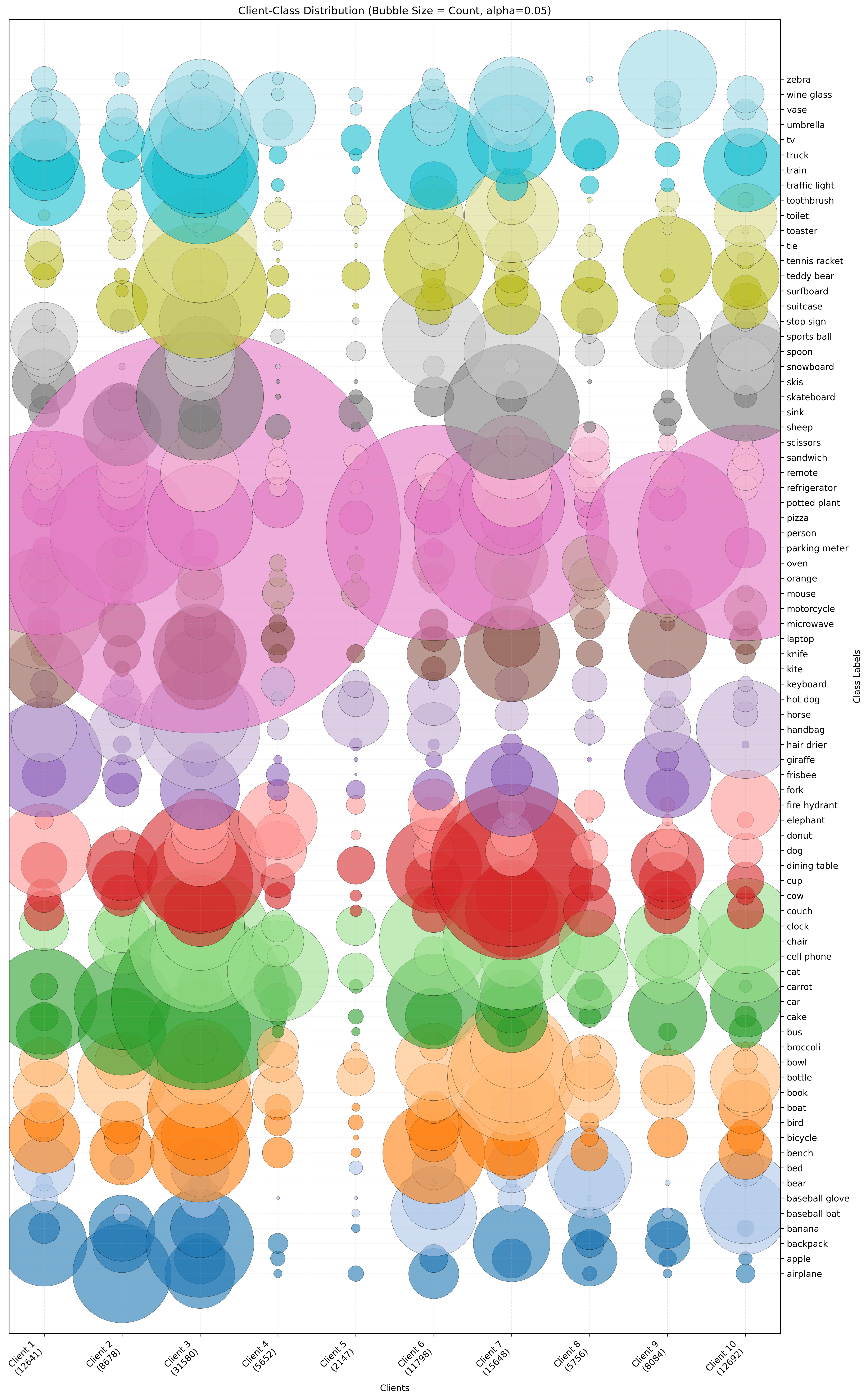}
            \caption{Per-client training data distribution under $\beta = 0.05$ and $\gamma = 0.75$.}
        \end{subfigure}
    \end{minipage}
    \caption{
    Distribution of data across local clients in the COCO~\cite{lin2014microsoft} experiments.
    The class presence ratio ($\gamma$) is set to 0.75 ($\leq$ 60 of 80 classes per client).
    Non-IID client distributions are simulated using the Dirichlet factor ($\beta$).
    }
    \label{fig:COCO_data_distribution}
\end{figure}

\noindent \textbf{MS COCO}~\cite{lin2014microsoft} is another widely used benchmark dataset in CV. 
It contains a large-scale collection of natural images, each annotated with one or more object categories selected from a predefined set of 80 classes. 
Compared with PASCAL VOC, MS COCO poses an even more challenging multi-label classification task due to its larger label space, more complex visual scenes, and denser object co-occurrence patterns. 
In particular, many images contain multiple objects with diverse scales, occlusions, and cluttered backgrounds, leading to substantial class imbalance, high intra-class variability, and frequent inter-class co-occurrence.
These categories include \textit{person}, \textit{bicycle}, \textit{car}, \textit{motorcycle}, \textit{airplane}, \textit{bus}, \textit{train}, \textit{truck}, \textit{boat}, \textit{traffic light}, \textit{fire hydrant}, \textit{stop sign}, \textit{parking meter}, \textit{bench}, \textit{bird}, \textit{cat}, \textit{dog}, \textit{horse}, \textit{sheep}, \textit{cow}, \textit{elephant}, \textit{bear}, \textit{zebra}, \textit{giraffe}, \textit{backpack}, \textit{umbrella}, \textit{handbag}, \textit{tie}, \textit{suitcase}, \textit{frisbee}, \textit{skis}, \textit{snowboard}, \textit{sports ball}, \textit{kite}, \textit{baseball bat}, \textit{baseball glove}, \textit{skateboard}, \textit{surfboard}, \textit{tennis racket}, \textit{bottle}, \textit{wine glass}, \textit{cup}, \textit{fork}, \textit{knife}, \textit{spoon}, \textit{bowl}, \textit{banana}, \textit{apple}, \textit{sandwich}, \textit{orange}, \textit{broccoli}, \textit{carrot}, \textit{hot dog}, \textit{pizza}, \textit{donut}, \textit{cake}, \textit{chair}, \textit{couch}, \textit{potted plant}, \textit{bed}, \textit{dining table}, \textit{toilet}, \textit{tv}, \textit{laptop}, \textit{mouse}, \textit{remote}, \textit{keyboard}, \textit{cell phone}, \textit{microwave}, \textit{oven}, \textit{toaster}, \textit{sink}, \textit{refrigerator}, \textit{book}, \textit{clock}, \textit{vase}, \textit{scissors}, \textit{teddy bear}, \textit{hair drier}, and \textit{toothbrush}, covering a broad range of everyday object categories.
In the label-skewed FL setting, these difficulties are further amplified by non-IID client distributions and inconsistent co-occurrence patterns across clients. 
We conduct experiments under the FL setting $\beta = 0.05, \gamma = 0.75$, and visualize the resulting class-wise client distributions in Figure~\ref{fig:COCO_data_distribution}.


\noindent \textbf{ChestX-ray14}~\cite{wang2017chestxray} is a large-scale medical imaging dataset widely used for automated thoracic disease classification. 
It contains 112,120 frontal-view chest X-ray images collected from 30,805 unique patients. 
Each image is annotated with one or more disease labels, making the dataset naturally suited to multi-label classification. 
These labels are extracted from the corresponding radiology reports using natural language processing techniques.
The dataset includes 14 disease categories: \textit{Atelectasis}, \textit{Cardiomegaly}, \textit{Consolidation}, \textit{Edema}, \textit{Effusion}, \textit{Emphysema}, \textit{Fibrosis}, \textit{Hernia}, \textit{Infiltration}, \textit{Mass}, \textit{Nodule}, \textit{Pleural Thickening}, \textit{Pneumonia}, and \textit{Pneumothorax}. 
In addition, a \textit{No Finding} label is used to indicate negative samples in which none of the 14 diseases are present.
Due to its large scale, real-world variability, and inherent label noise, ChestX-ray14 provides a challenging benchmark for developing and evaluating deep learning models in medical image analysis, particularly in multi-label and imbalanced scenarios.
We conduct experiments on this dataset under two label-skewed FL configurations: $\beta = 0.5$, $\gamma = 0.5$ and $\beta = 0.1$, $\gamma = 0.5$.
Since a significant portion of the dataset (57\% of the training data) is \textit{No Finding} samples, we distribute these samples evenly across all clients in both settings. 
This setup mimics realistic clinical scenarios where healthy cases are common, while disease cases are relatively rare and unequally distributed across institutions.
Figure~\ref{fig:Xray_data_distribution} illustrates the resulting class-wise data distributions across clients.

\section{Experiment (Additional)}
\label{sup_sec: experiment}
\subsection{Ablation Study (Additional)}
\label{sup_sub_sec: experiment}

\begin{figure}[t]
    \centering
    \begin{subfigure}[t]{0.32\linewidth}
        \includegraphics[width=\linewidth, trim={2.5cm 2.7cm 10cm 1cm}, clip]{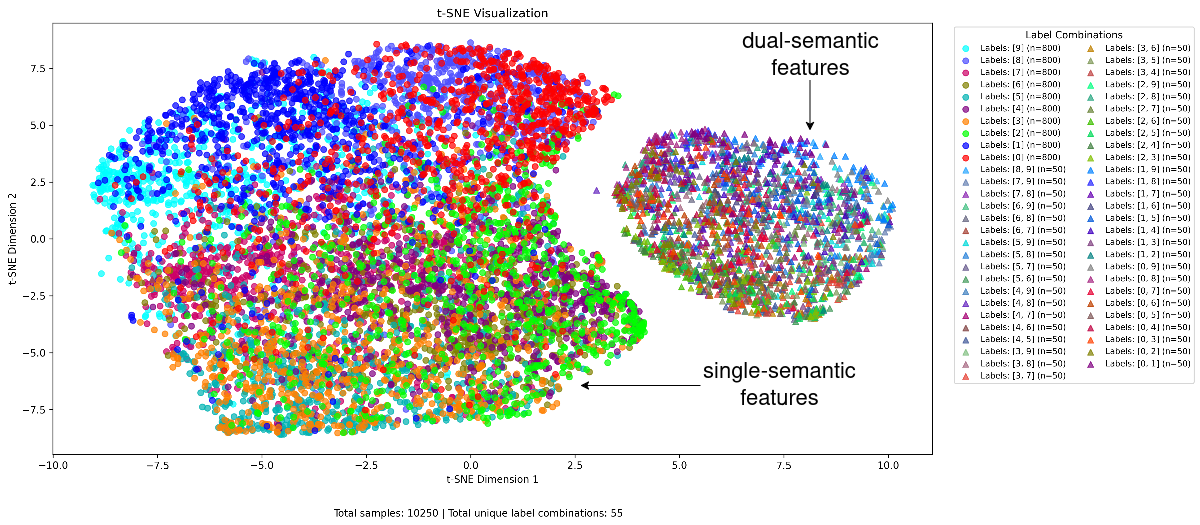}  % left bottom right top
        \subcaption{Learnable FC classifier}
    \end{subfigure}
    \begin{subfigure}[t]{0.32\linewidth}
        \includegraphics[width=\linewidth, trim={2.5cm 2.7cm 10cm 1cm}, clip]{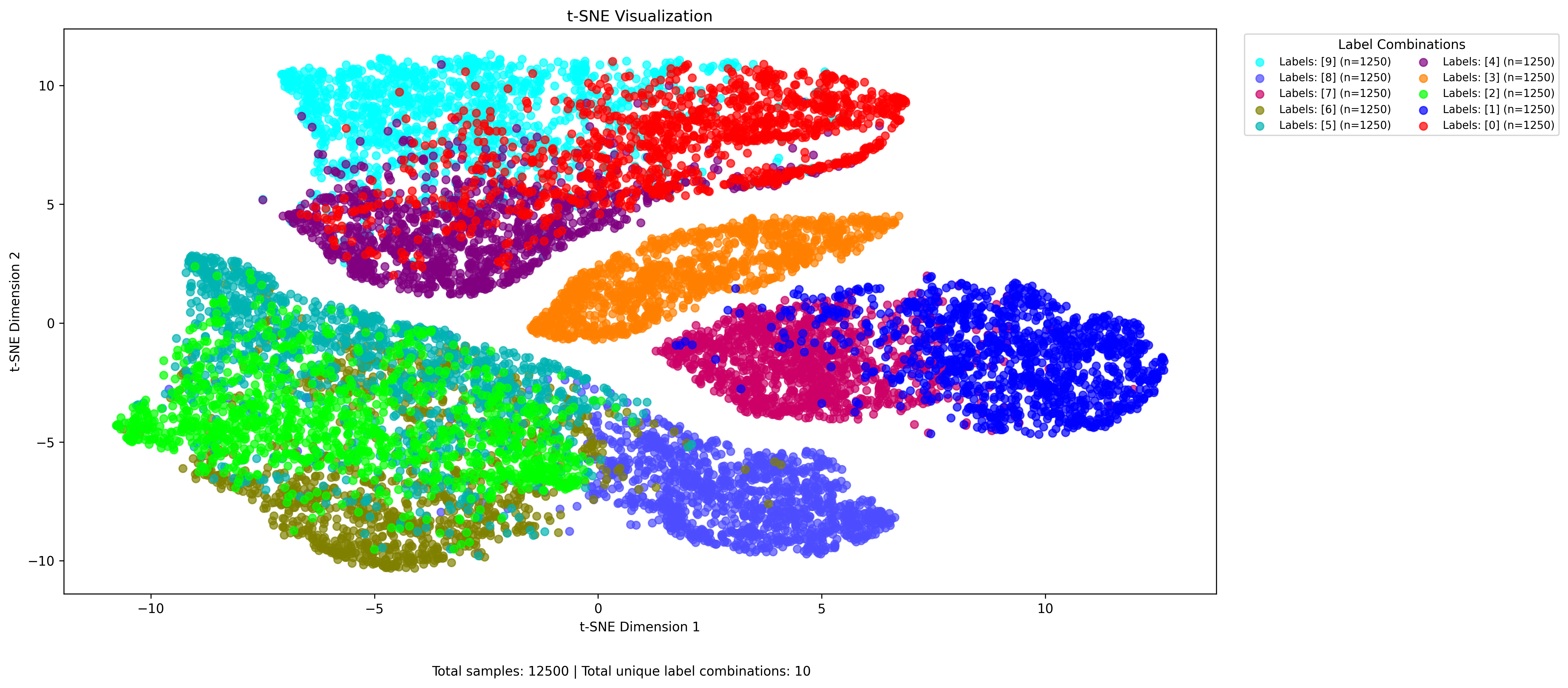}
        \subcaption{LADM + fixed ETF classifier}
    \end{subfigure}
    \begin{subfigure}[t]{0.32\linewidth}
        \includegraphics[width=\linewidth, trim={2.5cm 2.7cm 10cm 1cm}, clip]{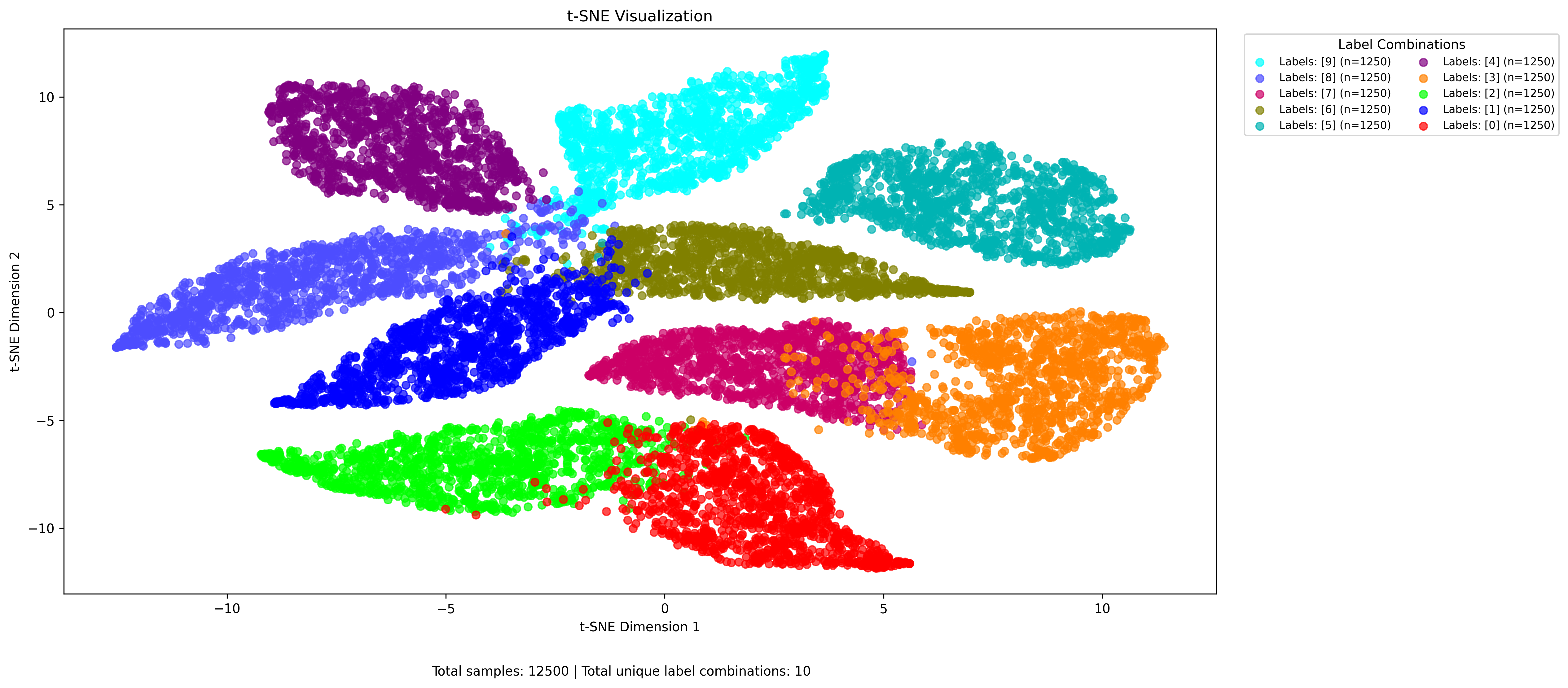}
        \subcaption{LADM + fixed ETF + Dual regularization}
    \end{subfigure}
    \caption{
    t-SNE visualisation of test data feature embeddings on the multi-label CIFAR-10 experiment with $\beta = 0.5$, $\gamma = 0.5$.
    Each colour represents a class.
    Observing from subfigure (a), without feature disentanglement (LADM), the model appears to rely on undesired information, such as the number of labels per sample, for clustering.
    }
    \label{fig: Ablation_CIFAR10_tsne}
\end{figure}

\begin{wrapfigure}{l}{0.6\linewidth}
    \centering
    \begin{subfigure}[t]{0.3\linewidth}
        \includegraphics[width=\linewidth, trim={0.4cm 0cm 4.5cm 0.785cm}, clip]{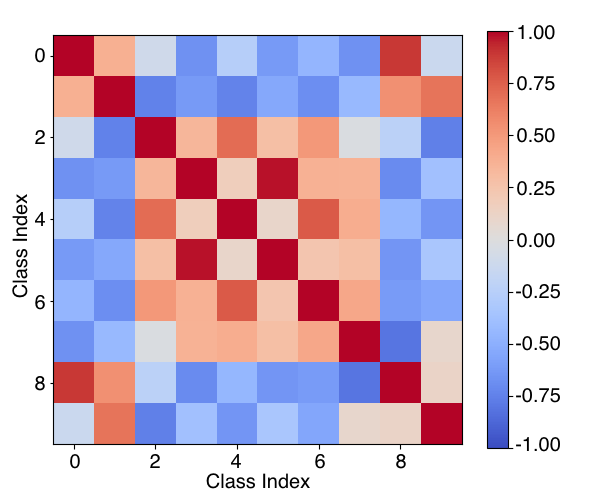}
        \subcaption{FC classifier}
    \end{subfigure}
    \begin{subfigure}[t]{0.3\linewidth}
        \includegraphics[width=\linewidth, trim={0.4cm 0cm 4.5cm 0.785cm}, clip]{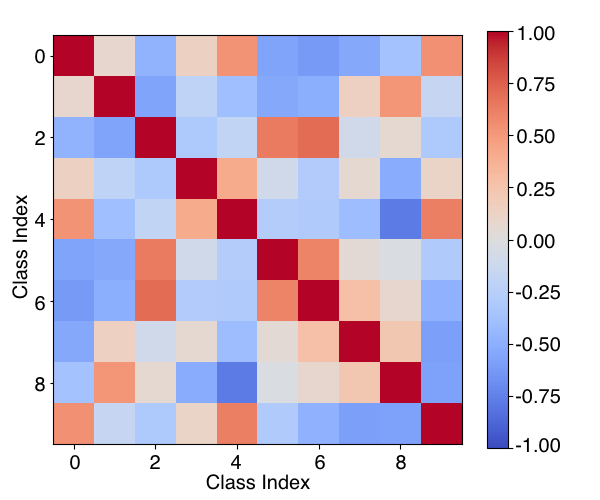}
        \subcaption{LADM + fixed ETF classifier}
    \end{subfigure}
    \begin{subfigure}[t]{0.358\linewidth}
        \includegraphics[width=\linewidth, trim={0.4cm 0cm 1.3cm 0.785cm}, clip]{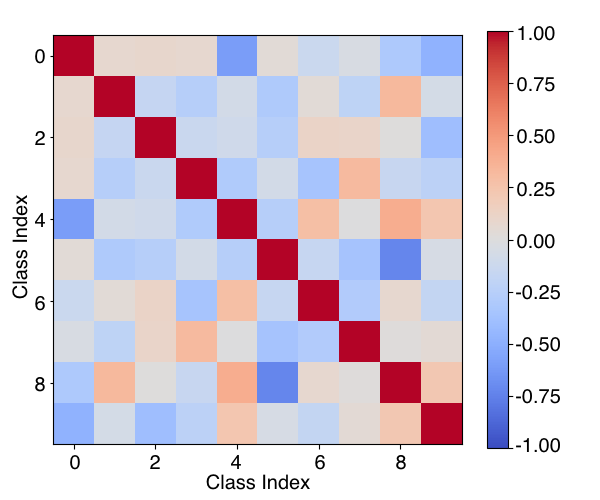}
        \subcaption{LADM + ETF + Dual regularization}
    \end{subfigure}
    \caption{
    Pair-wise cosine similarity between test data class-wise average feature prototypes on the multi-label CIFAR-10 dataset.
    }
    \label{fig: Ablation_CIFAR10_prototype}
    \vspace{-3mm}
\end{wrapfigure}

To complement the results reported in the main manuscript, we present an additional ablation study to evaluate the contribution of each component in our proposed method. 
The experiments are conducted on the multi-label CIFAR-10 dataset~\cite{krizhevsky2009learning}, where the client data distribution is configured using a Dirichlet concentration parameter of $\beta = 0.5$ and a class-presence ratio of $\gamma = 0.5$ (i.e., each client has access to at most 5 out of 10 classes).
The experimental results are summarized in Tables~\ref{tab: Ablation_CIFAR10} and \ref{tab: Ablation_LADM_CIFAR10}, and Figures~\ref{fig: Ablation_CIFAR10_tsne} and \ref{fig: Ablation_CIFAR10_prototype}.
As shown in Table~\ref{tab: Ablation_CIFAR10}, integrating LADM for class-specific feature extraction, along with the use of a predefined ETF classifier to encourage class-wise clustering alignment across clients in the latent feature space, leads to performance improvements of 3.38\% in class-wise AUC and 3.99\% in class-wise F1 score. 
Additional gains are achieved by introducing regularization terms that enhance the model’s discriminative capacity, resulting in a further class-wise increase of 1.88\% in class-wise AUC and 4.58\% in class-wise F1 score. 
Regarding LADM specifically, the results in Table~\ref{tab: Ablation_LADM_CIFAR10} demonstrate that using fixed, well-designed class-wise queries is generally more effective than learnable queries across most evaluation metrics.

\begin{table}[tbp]
    \centering
    \tiny
    \caption{
        Ablation study of the proposed method on the multi-label CIFAR-10 dataset~\cite{krizhevsky2009learning}. 
        $\beta$ is set to 0.5 and $\gamma$ is set to 0.5.
    }
    \vspace{-5pt}
    \renewcommand{\arraystretch}{1.5} % Adjust row separation
    \resizebox{.65\linewidth}{!}{
    \begin{tabular}{c c c c | c c c c}
    \hline 
    ETF Clf & LADM & $\mathcal{L}_{\text{Neg}}$ & $\mathcal{L}_{\text{Pos}}$ & macro-AUC & macro-F1 & micro-AUC & micro-F1 \\
    \hline
    & & & & 82.46 & 40.65 & 81.83 & 41.24 \\
    \checkmark &  & & & 83.44 & 15.29 & 81.75 & 16.77 \\
    \checkmark & \checkmark & & & 85.84 & 44.64 & 85.10 & 45.45 \\
    \checkmark & \checkmark & \checkmark & & 87.72 & 48.22 & 86.16 & 48.93 \\
    \checkmark & \checkmark & & \checkmark & 86.78 & 44.46 & 86.08 & 45.28 \\
    \checkmark & \checkmark & \checkmark & \checkmark & \textbf{87.72} & \textbf{49.22} & \textbf{87.13} & \textbf{49.60} \\
    \hline
    \end{tabular}}
    \label{tab: Ablation_CIFAR10}
    
% ================================================================================================
\vspace{1em}
    \centering
    \scriptsize
    \caption{
        Ablation study of the class-wise feature extraction block - LADM on the multi-label CIFAR-10 dataset. 
        % with $\beta = 0.1$, $\gamma = 0.71$.
    }
    \resizebox{.65\linewidth}{!}{
    \renewcommand{\arraystretch}{1.2} % Adjust row separation
    \begin{tabular}{c c | c c c c }
    \hline 
    query type & query init & macro-AUC & macro-F1 & micro-AUC & micro-F1 \\
    \hline
    learnable & random & 84.15 & 38.10 & 83.70 & 39.30 \\
    learnable & ETF & \textbf{86.26} & 43.61 & \textbf{85.41} & 43.87 \\
    fixed & ETF & 85.84 & \textbf{44.64} & 85.10 & \textbf{45.45} \\
    \hline
    \end{tabular}}
    \label{tab: Ablation_LADM_CIFAR10}
\end{table}

To further analyse model behaviour, we visualize the latent feature distributions of the test set using t-SNE under different architectural and training configurations.
As shown in Figure~\ref{fig: Ablation_CIFAR10_tsne}, incorporating LADM for feature disentanglement, along with a predefined ETF classifier to regulate feature distribution across clients, 
enables the model to focus on semantic content rather than irrelevant factors such as the number of labels present in each sample.
Furthermore, the addition of regularization terms during training leads to more compact and semantically coherent clusters, while also reducing the similarity among class-wise average prototypes. 
As illustrated in Figure~\ref{fig: Ablation_CIFAR10_prototype}, the pairwise cosine similarity between class-wise average features decreases with the inclusion of LADM and the ETF classifier, and is further reduced by the added regularization terms, indicating enhanced inter-class separability and stronger discriminative capability.

\end{document}